\documentclass[11pt]{article}

\usepackage{acl}

\usepackage{times}
\usepackage{latexsym}

\usepackage[T1]{fontenc}

\usepackage[utf8]{inputenc}

\usepackage{microtype}

\usepackage{inconsolata}

\usepackage{graphicx}

\usepackage{mathtools}

\usepackage{hyperref}
\usepackage{url}
\usepackage{amssymb}
\usepackage{mathtools}
\usepackage{amsmath}
\usepackage{amsthm}
\usepackage{algorithm}
\usepackage{algorithmic}

\usepackage{xcolor}
\usepackage{colortbl}
\usepackage{wrapfig} 
\usepackage{tabularx}
\usepackage{booktabs}
\usepackage{bm}

\newcommand{\pii}{\pi_\theta}                    
\newcommand{\piiold}{\pi_{\theta_{\text{old}}}}  

\newcommand{\piit}[1]{\pi_{\theta^{(#1)}}}       

\newcommand{\EE}{\mathbb{E}}                 
\newcommand{\grad}{\nabla}                    

\definecolor{verylightgreen1}{RGB}{240,255,240} 
\definecolor{mygreen1}{RGB}{220,255,220}   
\definecolor{mygreen2}{RGB}{190,240,190}   

\newcommand{\da}{\textsuperscript{$\dagger$}} 
\newcommand{\di}{\textsuperscript{$\diamond$}} 

\title{Data-Efficient RLVR via Off-Policy Influence Guidance}

\author{Erle Zhu*\da, Dazhi Jiang*\da\thanks{* denotes equal contribution. Work was done when EZ \& DJ interned at Z.ai. }, Yuan Wang\da, Xujun Li\da, Jiale Cheng\da, 
Yuxian Gu\da\\
\textbf{Yilin Niu\di, Aohan Zeng\di, Jie Tang\di, Minlie Huang\da,  Hongning Wang\da}\\
\da~CoAI Group, Tsinghua University\\
\di~Z. AI \\
\texttt{\{zel24\}@mails.tsinghua.edu.cn, hw-ai@tsinghua.edu.cn} \\
}

\begin{document}
\maketitle
\begin{abstract}
Data selection is a critical aspect of Reinforcement Learning with Verifiable Rewards (RLVR) for enhancing the reasoning capabilities of large language models (LLMs).
Current data selection methods are largely heuristic-based, lacking theoretical guarantees and generalizability. 
This work proposes a theoretically-grounded approach using influence functions to estimate the contribution of each data point to the learning objective. 
To overcome the prohibitive computational cost of policy rollouts required for online influence estimation, we introduce an off-policy influence estimation method that efficiently approximates data influence using pre-collected offline trajectories.
Furthermore, to manage the high-dimensional gradients of LLMs, we employ sparse random projection to reduce dimensionality and improve storage and computation efficiency.
Leveraging these techniques, we develop \textbf{C}urriculum \textbf{R}L with \textbf{O}ff-\textbf{P}olicy \text{I}nfluence guidance (\textbf{CROPI}), a multi-stage RL framework that iteratively selects the most influential data for the current policy. 
Experiments on models up to 7B parameters demonstrate that CROPI significantly accelerates training. 
On a 1.5B model, it achieves a $2.66\times$ step-level acceleration while using only 10\% of the data per stage compared to full-dataset training. 
Our results highlight the substantial potential of influence-based data selection for efficient RLVR.
Code is available at \url{https://github.com/thu-coai/CROPI}.

\end{abstract}

\section{Introduction}

The advent of highly capable reasoning models, such as OpenAI o series models \citep{openai2024reasoning} and DeepSeek R1 \citep{guo2025deepseek}, have established Reinforcement Learning with Verifiable Rewards (RLVR) as a key step for enhancing the reasoning capabilities of large language models (LLMs). 
Data quality is critical to model performance, making data selection for RLVR a key research area.
Existing data selection methods for RLVR \citep{wang2025reinforcement,bae2025online,li2025limr,zhao2025ufo, sun2025improving} are primarily heuristic-based, often focusing on metrics like difficulty or uncertainty; but such metrics lack theoretical performance guarantees and exhibit poor generalizability across different scenarios.

In this work, we propose using influence functions \citep{hampel1974influence, koh2017understanding} for RL data selection. 
This method approximates the contribution of a given data point to the learning objective via variational analysis in calculus. 
Compared to heuristic-based approaches, influence functions offer stronger theoretical guarantees and provide more fine-grained information about data effect. 

However, applying influence functions to RLVR for large-scale models faces a significant barrier. 
Unlike supervised learning (e.g., pre-training or supervised fine-tuning) where supervision is readily available, RL supervision must be generated through policy rollouts \citep{grosse2023studying, xia2024less}. 
For LLMs, these rollouts are computationally expensive, which makes the online estimation of data influence prohibitively difficult. 
To address this challenge, we propose a method to estimate data gradients using pre-collected offline trajectories. 
This approach allows for the efficient evaluation of a data point's influence on the online policy without requiring new, costly rollouts. 
Furthermore, to overcome the challenges of storing and computing the high-dimensional gradients typical of LLMs, we employ sparse random projection. 
This technique maps the gradients to a lower-dimensional space, thereby improving storage efficiency and mitigating numerical noise.

Leveraging off-policy influence estimation, we develop a curriculum-based reinforcement learning framework named \textbf{C}urriculum \textbf{R}L with \textbf{O}ff-\textbf{P}olicy \textbf{I}nfluence guidance (\textbf{CROPI}).
CROPI segments the RL training process into multiple stages. 
In each stage, it selects the subset of data with the highest estimated influence on the current policy checkpoint for subsequent training. 
We demonstrate CROPI's effectiveness through experiments on models of varying sizes (1.5B to 7B) and context lengths. 
On the 1.5B model, CROPI achieves a $\bm{2.66\times}$ step-level acceleration compared to full-dataset training, while using only \textbf{10\%} of the data in each stage. 
This result highlights the substantial potential of influence-based data selection for online RLVR.

In summary, our contributions are as follows:
\begin{itemize}
\item We introduce Off-Policy Influence Estimation, a theoretically-grounded and rollout-free method to quantify the influence of individual data points on an online policy, eliminating the need for real-time sampling.
\item To efficiently handle the high-dimensional gradients of LLMs, we employ Sparse Random Projection for dimensionality reduction. We empirically demonstrate that applying dropout prior to this projection mitigates numerical noise and enhances computational efficiency while preserving inner products.
\item We propose CROPI, a curriculum reinforcement learning framework that leverages our influence estimation method for multi-stage data selection. Our experiments show that CROPI substantially outperforms both full-dataset training and alternative data selection baselines.
\end{itemize}

\section{Preliminaries}
\label{sec:pre}

\begin{figure*}[ht]
    \centering
    \includegraphics[width=\textwidth]{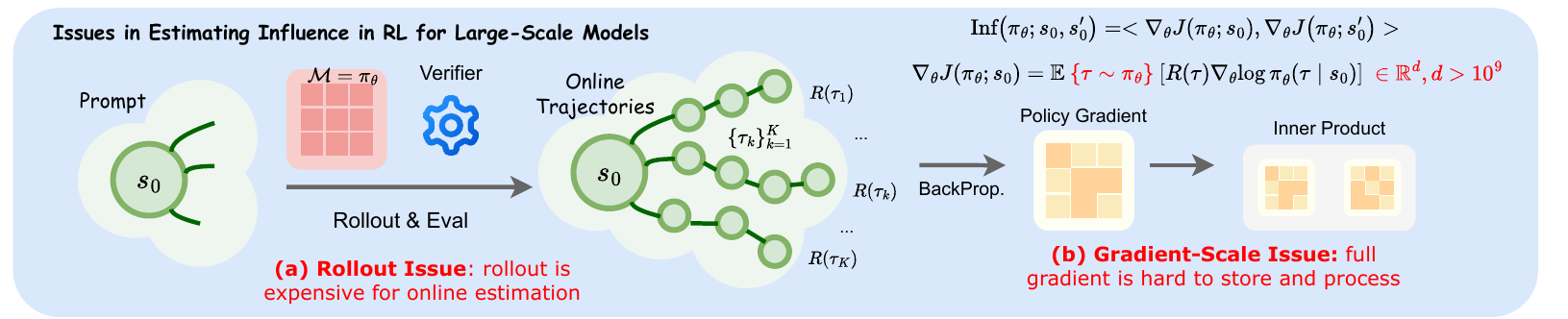}
    \caption{Practical issues in computing influence for data point in RL training process for large-scale models. 
    }
    \label{fig:opi_motiv}
\end{figure*}


\noindent\textbf{Reinforcement Learning with Verifiable Rewards (RLVR).} Using the language of reinforcement learning (RL), the reasoning process of LLMs can be modeled as a Markov Decision Process $\mathcal{M} = (\mathcal{S}, \mathcal{A}, P, r, \gamma)$ \citep{sutton1999reinforcement}. 
Let $\mathcal{V}$ denote the vocabulary, with each generated token $x \in \mathcal{V}$. The state space $\mathcal{S} = \mathcal{V}^*$ consists of all possible token sequences, and actions correspond to generating the next token, such that $\mathcal{A} = \mathcal{V}$. 
The autoregressive generation process produces $x_t$ at each step $t$ given the prefix $s_{t}$, with a deterministic state transition: $s_{t+1} = s_t || x_{t}$, where $||$ denotes concatenation. 
The reward function is outcome-based: $r_t = 0$ for $t < T$, and $r_T = R_{\text{correct}}(y)$, where $y$ is the solution extracted from the final sequence $s_T$, and $R_{\text{correct}}(y) \in \{0, 1\}$ is a deterministic correctness indicator. 
A reasoning trajectory is defined as $\tau = \{(s_0, x_0), ..., (s_{T-1}, x_{T-1})\}$, and its return is $R(\tau) = \sum_{t=1}^{T}r_t = R_{\text{correct}}(y)$  (with a discount factor $\gamma=1$). 

With Deepseek-R1 \citep{guo2025deepseek} emerging as the most powerful open-source reasoning model, its RL algorithm GRPO \citep{shao2024deepseekmath}, has become the mainstream approach for RLVR. 
This algorithm builds upon PPO \citep{schulman2017proximal}, but estimates the advantage using group-normalized returns, thereby eliminating the overhead of the critic model required in PPO. 
The optimization objective is as follows (ignoring the clipping \& KL term): 
\begin{align}
 J(\theta) &= \mathbb E_{s_0\sim q(\cdot),\{\tau_k\}_{k=1}^K \sim \piiold} \sum_{k=1}^{K}  \frac{1}{|T_k|}
 \sum_{t=0}^{T_k-1} \rho_{k,t}^{\pii} \widehat{A}_{k,t} \label{eq:grpo}
\end{align}
where $ \rho_{k,t}^{\pii} = \frac{\pii(x_{k,t}|s_{k,t})}{\piiold(x_{k,t}|s_{k,t})} $ and $ \widehat{A}_{k,t}= \frac{R(\tau_k)-\widehat{\mathbb E}_{\piiold} [R(\tau)]}{\widehat{\sigma}_{\piiold}[R(\tau)]} $. 
$\widehat{\EE}$ and $\widehat{\sigma}$ denote the empirical mean and standard deviation of the returns of trajectories $\{\tau_k\}_{k=1}^K$ sampled from $\piiold$ given query $s_0$.



\noindent{\textbf{Influence Function.}}
Influence functions \citep{hampel1974influence, koh2017understanding} offer a gradient-based approach for data attribution, derived based on the variational analysis of the objective function. 
Briefly, given an objective function $J$ to be maximized and a collection of $N$ data points $z_i $, suppose the model parameters are updated from $ \theta_0 $ to $ \theta_T $. Our goal is to estimate the contribution (or influence) of each individual data point to the change in the objective function: 
$J(\theta_T)=J(\theta_0)+\sum_{i=1}^N\text{Influence}(z_i)$. 
Owing to strong theoretical guarantees and empirical success, influence function and its variants have been widely applied in pre-training and supervised fine-tuning stages of LLMs \cite{grosse2023studying, gu2024data, xia2024less, wang2024greats} for data attribution and selection. 
However, how to leverage influence functions for data selection in RLVR for LLMs remains an open question.

\section{Influence Estimation in RLVR}
\label{sec:opi}

Following the first-order influence function formula \citep{pruthi2020estimating}, in the RLVR context, given a training prompt $s_0$ and a test query $s_0'$, we use the inner product of the policy gradients of $s_0$ and $s_0'$ to measure the influence of $s_0$ on the policy's performance on $s_0'$:
\begin{align}
\label{eq:inf} \text{Inf}(\pii; s_0, s_0') &\coloneqq \langle \nabla_\theta J(\theta; s_0 ), \nabla_\theta J(\theta; s_0') \rangle.  \\
\label{eq:pg} \text{where}~~\nabla_\theta J(\theta;s_0) &= \mathbb E_{\tau\sim\pi_\theta} [R(\tau)\nabla_\theta \log \pii (\tau|s_0)] 
\end{align}
$\nabla J(\theta;s_0)$ denotes policy gradient for initial state $s_0$, we use a vanilla version for simplicity. 
We give a short derivation for Equation \ref{eq:inf} in Appendix \ref{sec:inf_theory}. 
Although the notion of influence function possesses favorable theoretical for the problem of data selection, its practical estimation for RL algorithms still faces two main challenges, namely the \textbf{Rollout Issue} and the \textbf{Gradient-Scale Issue} as illustrated in Figure \ref{fig:opi_motiv}.

\noindent \textbf{Rollout Issue.} 
Unlike supervised learning whose training dynamics are predominantly shaped by labeled data and optimization algorithms \citep{xia2024less, gu2024data}, RL involves online sampling, making its training process more dynamic and less predictable. 
Consequently, global data selection strategies \citep{zhao2025ufo, wang2025reinforcement} can hardly capture the evolving characteristics of policy learning in RLVR. 
Therefore, we seek to dynamically evaluate the influence of each data point $s_0$ conditioned on the current policy $\pii$, thereby enabling effective online utility estimation. 
However, accurate influence estimation typically requires 
computing the policy gradient for each $s_0$, which demands rollouts over multiple trajectories (Eq. \eqref{eq:pg}). 
The substantial computational costs and latency associated with these rollouts present a significant barrier to real-time influence estimation in LLMs.
In our 1.5B setting with batch size 128 and maximum response length 8192, one rollout step takes 319.80s on average, while one forward+backward step takes only 35.91s, making rollout about 8.91x more expensive.
This gap is also consistent with the underlying hardware pattern: forward+backward is mostly compute-bound, while autoregressive decoding is largely memory-bandwidth-bound.
Therefore, a rollout-free utility estimator is not merely a convenience but a key requirement for making online data selection practical; we provide additional timing details in Appendix \ref{sec:appendix_rollout_cost}. 

\noindent \textbf{Gradient-Scale Issue.}
Moreover, the high dimensionality of gradients in large-scale models poses additional storage and computational challenges. 
For instance, \citet{xia2024less} mitigate this issue through random projection, leveraging the Johnson-Lindenstrauss Lemma \citep{johnson1984extensions} to efficiently preserve inner products with high probability. 
However, while \citet{xia2024less} utilize LoRA \citep{hu2022lora} to reduce raw gradient dimensionality during SFT, we consider full-parameter training for better performance assurance in RLVR, resulting in massive raw gradients and high projection overhead. 

To address these two issues, we propose an off-policy gradient estimation technique to eliminate the reliance on costly rollouts, and employ sparse random projection methods to achieve scalable and efficient gradient storage and computation. 
This dual approach facilitates practical and effective online influence estimation within RLVR, making it suitable for large-scale model training scenarios.

\subsection{Off-Policy Gradient for Rollout-Free Estimation}
\label{sec:op_grad}

To address the \textbf{Rollout Issue}, following the ideas in offline RL \cite{levine2020offlinereinforcementlearningtutorial} which use offline trajectories to perform RL algorithms, we use offline trajectories generated by behavior policy $\beta$ to compute the gradient for policy $\pii$ and use the off-policy gradient to estimate the influence of the prompt $s_0$, as shown on Figure \ref{fig:cropi_framework} (c).
Given a data point $s_0$, RL policy $\pi_\theta$ trained with group-norm advantage estimator, behavior policy $\beta$ and $K$ offline trajectories ${\{\tau_k\}_{k=1}^{K} \sim \beta(\cdot|s_0)}$ sampled from behavior policy $\beta$ conditioned on $s_0$. 
If $\pi_\theta$ and $\beta$ are KL-constrained, we can approximate the on-policy gradient with an off-policy estimator:
\vspace{-0.5em}
\begin{align}
    \widehat{g}_\beta (\theta, s_0, \{\tau_k\}_{k=1}^K) &\approx  \frac{1}{K} \sum_{k=1}^{K}  \frac{1}{|\tau_k|} \sum_{t=0}^{|\tau_k|-1} \nabla_\theta \rho_{k,t}^{\pii}\widehat{A}^{\beta}_{k,t} \label{eq:op_grad}
\end{align}
\vspace{-0.3em}
Where $\rho^{\pii}_{k,t} = \frac{\pii(x_{k,t}|s_{k,t})}{\beta(x_{k,t}|s_{k,t})}$
and $\widehat{A}_{k,t}^{\beta} = \frac{ R(\tau_k) - \widehat{\EE}_{\beta} [R(\tau) ]}{\widehat\sigma_\beta [R(\tau)]}$. 
This gradient is derived from TRPO method \citep{schulman2015trust}, please refer to Appendix \ref{sec:opi_theory} for more details.
This gradient is equivalent to removing the clipping operation from the GRPO objective and replacing $\piiold$ with $\beta$ in the gradient computation.

In this work, we select $\beta = \pi_{\theta_0}$.
Since there is a KL-Term in online RL objective \citep{shao2024deepseekmath, ouyang2022training}, the KL-distance of $\pi_{\theta_0}$ and $\pii$ is usually constrained, making our off-policy gradient estimator relatively accurate. 
In our setting, we sample multiple trajectories by $\pi_{\theta_0}$ for every prompt in the dataset before training, 
resulting in $\mathcal D = \{s_0^{(i)}, \{\tau_k^{(i)}\}_{k=1}^{K} \}_{i=1}^{N}$, which is a necessary process for all existing data selection methods in RLVR. 
We denote $\widehat{g}_\beta (\theta, s_0, \{\tau_k\}_{k=1}^K)$ as $\widehat{g}_\beta (\theta, s_0)$ for simplicity.

Based on off-policy gradient $\widehat{g}_\beta (\theta, s_0)$, we can measure the influence between a training data $s_0$ and validation data $s_0'$.
Following Equation \ref{eq:inf}, our off-policy influence estimation can be formulated as $\widehat{\text{Inf}}_\beta(\pi_\theta; s_0, s_0')$: 
\begin{align}
\label{eq:opi}
    \widehat{\text{Inf}}_\beta(\pi_\theta; s_0, s_0') &= \widehat{g}_\beta (\theta, s_0 )^\top\widehat{g}_\beta (\theta, s_0')
\end{align}
The experimental results indicate that the off-policy gradient can approximate the on-policy gradient to a certain extent; please refer to Appendix \ref{sec:off_policy_estimator} for further details.

\subsection{Sparse Random Projection for Full Gradient Compression}
\label{sec:spar_proj}

To address the \textbf{Gradient-Scale Issue}, we propose sparse random projection, a method where a subset of gradient dimensions is randomly omitted before the projection is performed.
Let the gradient be denoted by $g \in \mathbb{R}^d$, where $d$ is its dimensionality. 
A standard random projection uses a matrix $\mathcal{P} \in \mathbb{R}^{k \times d}$, where $k \ll d$ and each element $\mathcal{P}_{i,j}$ is sampled from a standard normal distribution, $\mathcal{N}(0,1)$. 
Our method first samples a random set of indices $S \subset \{1, \dots, d\}$. 
It then constructs a sparse random projection matrix, $\mathcal{P}_{\text{sparse}} \in \mathbb{R}^{k \times d}$, where the columns of $\mathcal{P}_{\text{sparse}}$ corresponding to indices in $S$ are sampled from $\mathcal{N}(0,1)$, while all other columns are zero vectors, i.e. $\mathcal{P}_{\text{sparse}}[i,j]=\epsilon_{i,j} \mathbb{I}_{j \in S},\epsilon_{i,j} \in \mathcal N(0,1)$, where $\mathbb{I}$ denotes indicator function.

This process is equivalent to first selecting a random subset of the gradient's dimensions and then applying a smaller projection. This can be expressed as $\mathcal{P}_{\text{sparse}} g = \mathcal{P}_{\text{sparse}}[:, S] g[S]$, where $g[S]$ is the subvector of $g$ with a subset of elements indexed by $S$, and $\mathcal{P}_{\text{sparse}}[:, S]$ is the submatrix of $\mathcal{P}_{\text{sparse}}$ formed by the columns indexed by $S$. We define the number of selected dimensions as $r_s = |S|$.
Our empirical results show that the random dropout of the gradient dimensions before projection achieves efficient and accurate rank preservation in inner product compared to directly conducting random projection. Please refer to Section \ref{sec:ana_random_projection} for more details.

\subsection{Final Solution} 

In order to eliminate the bias of the gradient norm caused by different lengths and pass rates, following \citet{xia2024less}, we normalize the gradient features before taking the inner product, which is equivalent to computing cosine similarity. 

Combining the off-policy gradient estimation and sparse random projection, denote $\widetilde{g}_\beta = \mathcal P_{\text{\small sparse}} ~\widehat{g}_\beta$, the practical computation of off-policy influence $\widehat{\text{Inf}}_\beta(\pi_\theta; s_0, s_0')$ can be formulated  as $\widetilde{\text{Inf}}_\beta(\pi_\theta; s_0, s_0')$: 
\begin{align}
 \widetilde{\text{Inf}}_\beta(\pi_\theta; s_0, s_0') &\coloneqq \texttt{cossim} \left( \widetilde{g}_\beta (\theta, s_0 ),~\widetilde{g}_\beta (\theta, s_0') \right)
\end{align}
We call this computation $\widetilde{\text{Inf}}_\beta(\pi_\theta; s_0, s_0')$ the Practical Off-Policy Influence estimation (POPI) for simplicity.

\section{Curriculum RL with Off-Policy Influence Guidance}
\label{sec:cropi}

\begin{figure*}[ht]
    \centering
    \includegraphics[width=\textwidth]{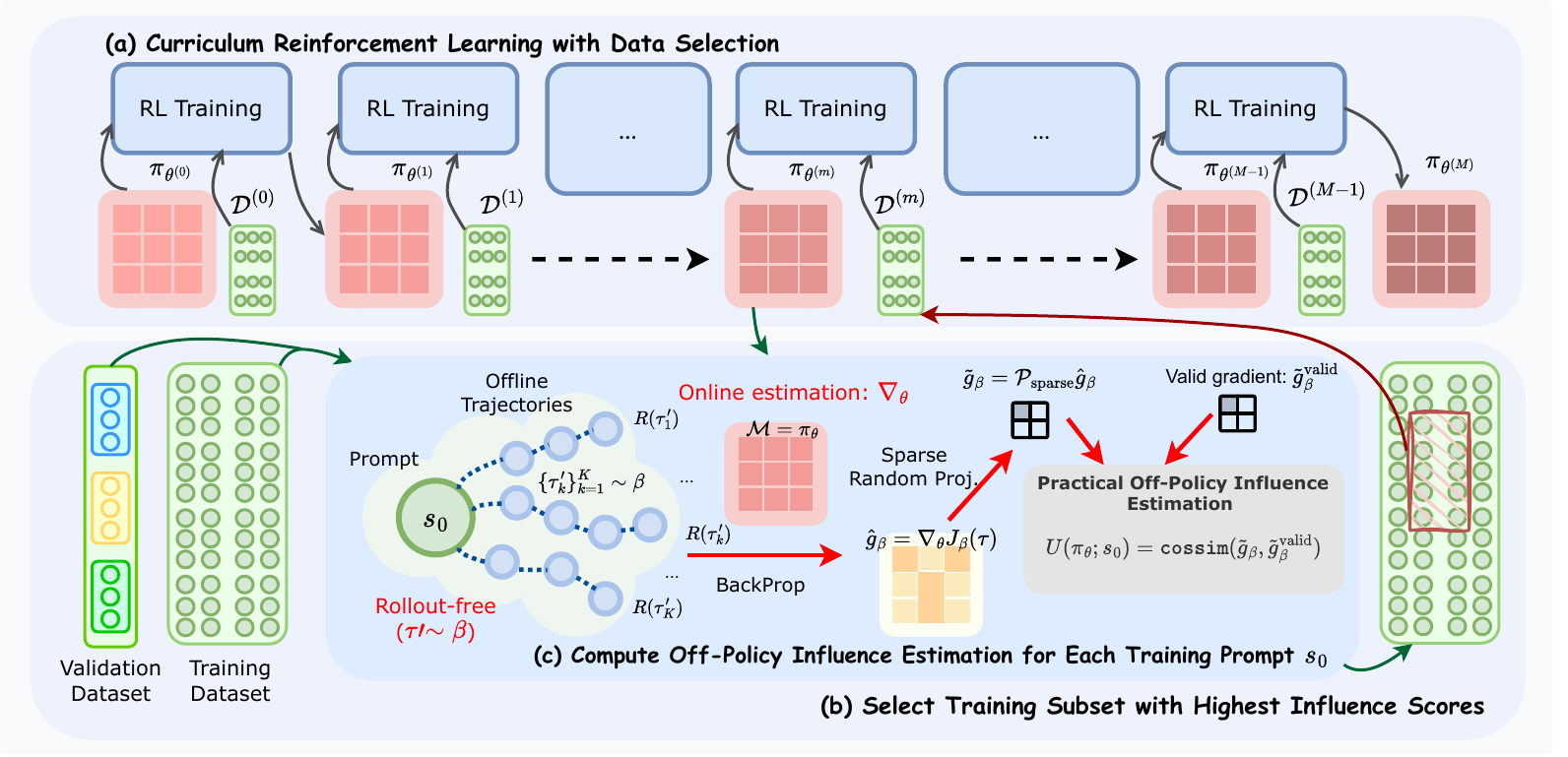}
    \caption{The schematic of our proposed framework \textbf{C}urriculum \textbf{R}L with \textbf{O}ff-\textbf{P}olicy \textbf{I}nfluence Guidance (\textbf{CROPI}).}
    \label{fig:cropi_framework}
\end{figure*}

In this chapter, we will illustrate how to use POPI estimator to select influential data for efficient RL training. 

\subsection{Data Selection with POPI}
\label{sec:sel_popi}

To measure the influence of training prompt $s_0$ to a specific validation set, we need to compute POPI over a batch of validation prompts. 
For a validation set $\mathcal D_{\text{val}} = \{s_0'^{(i)}\}_{i=1}^{N_{\text{val}}}$,
we denote the gradient feature of a validation set as the average gradient feature over all validation data points: $\widetilde{g}_\beta (\theta, \mathcal D_{\text{val}}) \coloneqq \sum_{i=1}^{N_{\text{val}}} \widetilde{g}_\beta (\theta,s_0'^{(i)}) $. Then the POPI of training prompt $s_0$ to the validation set $\mathcal D_{\text{val}}$ is defined as: 
\begin{align}
    \nonumber \widetilde{\text{Inf}}_\beta(\pi_\theta; s_0, \mathcal D_{\text{val}}) &\coloneqq \texttt{cossim} ( \widetilde{g}_\beta (\theta,s_0),~  \widetilde{g}_\beta (\theta, \mathcal D_{\text{val}}) )
\end{align}

For multiple validation sets, we need to consider the effectiveness over different validation sets. 
We use Reciprocal Rank Fusion (RRF) \citep{cormack2009reciprocal}, which is widely used in information retrieval, to combine the POPI scores over different validation sets. 
Denote the training set as $\mathcal D_{\text{tr}}$, number of $V$ validation sets as $\mathcal D_{\text{val}, j}, j=1,2,...V$, the rank of POPI score of training data $s_0$ to $j$-th validation set over the whole training set as $\text{r}_j(s_0) = \text{rank}_{\text{POPI}}(\pii; s_0, \mathcal D_{\text{val}, j})$, then the fused rank score is defined as: 
\begin{align}
    U_{\text{POPI-R}}(\pii; s_0)=\sum_{j=1}^{V} \frac {1} {\text{r}_j(s_0)} \label{eq:popi_r}
\end{align}
For any validation set, a higher rank (i.e. a small rank value) contribute to obtaining a higher RRF score.
In practice, given a RL policy checkpoint $\pii$, we select a subset $\mathcal{D}_{\text{sel}}$ from the whole training set using $U_{\text{POPI-R}}$ scores: 
\begin{align}
\mathcal{D}_{\text{sel}} = \underset{\mathcal{S} \subset \mathcal{D}, |\mathcal{S}| = \lfloor \alpha |\mathcal{D}| \rfloor}{\text{argmax}} \sum_{s_0 \in S} U_{\text{POPI-R}}(\pii; s_0) \label{eq:sel}
\end{align}

\subsection{Curriculum RL with Data Selection}
\label{sec:curriculum}
To balance long-term planning with dynamic selection, we adopt a phase-level data selection strategy with POPI, \textbf{C}urriculum \textbf{R}L with \textbf{O}ff-\textbf{P}olicy \textbf{I}nfluence Guidance (\textbf{CROPI}), which is illustrated on Figure \ref{fig:cropi_framework}. 
CROPI is an iterative curriculum-based RL framework designed to progressively filter and focus training on the most influential data points. 

At the start of each phase $m$, the current policy $\pi_{\theta^{(m)}}$ is evaluated on all training instances, yielding POPI-R scores $U_{\text{POPI-R}}(\pi_{\theta^{(m)}}; s_0^{(i)})$ (Eq. \ref{eq:popi_r}) for each $s_0^{(i)} \in \mathcal{D}_{\text{tr}}$. 
The subset $\mathcal{D}^{(m)}$ comprises the highest scoring samples, specifically $\lfloor \alpha |\mathcal{D}_{\text{tr}}| \rfloor$ instances, where $\alpha$ is the selection ratio. 
This targeted selection implements a dynamic curriculum, focusing subsequent policy optimization on the most impactful data points.
The policy is then refined on the selected subset using the GRPO algorithm over $E$ steps, producing an improved policy $\pi_{\theta^{(m+1)}}$ for the next phase. 
This iterative procedure is repeated for $M$ phases, resulting in a final policy $\pi_{\theta^{(M)}}$ output by CROPI.
We formulate this process in Algorithm \ref{alg:cropi} in Appendix \ref{sec:alg_cropi}.
This curriculum-based filtering and optimization scheme encourages the policy to focus on examples with the greatest potential benefit, thus accelerating learning process of RLVR. 

\section{Experiments}

\begin{table*}[t]
\centering
\footnotesize
\caption{
Evaluation results of CROPI and other baseline methods across various math datasets. 
CROPI consistently outperforms existing data selection approaches, achieving state-of-the-art performance on target tasks at different training steps and under various experimental settings. 
The best results at each training step are highlighted in \textbf{bold}. All \texttt{Acc.@step} values are computed from moving-average-smoothed evaluation curves with a window size of $5$.
}
\label{tab:main_result}

\begin{tabular*}{\textwidth}{@{\extracolsep{\fill}} p{3cm}p{0.8cm}p{0.8cm}p{0.8cm}p{1cm}p{1cm}p{0.8cm}p{1.7cm}p{1.5cm}}
\toprule

\multicolumn{1}{c}{$\uparrow $\textbf{Acc.(\%)}}  & {GSM8K}   & {MATH} & {Gaokao.}
&{AMC23}   & {Olympiad.} & {AIME24}  
  & \textbf{Targeted(Avg.)}   & {Untar.(Avg.)} \\

\midrule 
  
\rowcolor{verylightgreen1}{Qwen2.5-1.5B-Instruct}  & 72.99 & 55.48 & 46.43 & 28.13  & 24.27 & 4.00 & 64.24 & 25.71 \\
\midrule

\rowcolor{mygreen1}{\scriptsize~+~Full Dataset(GRPO)@500} & 78.01 & 58.07 & 47.34 & 32.50 & 26.32 & 1.67 & 68.04 & 26.96 \\
\rowcolor{mygreen2}{\scriptsize~+~Full Dataset(GRPO)@1k} & 79.30 & 59.54 & 44.09 & 31.25 & 26.23 & 6.67 & 69.42 & 27.06 \\
\rowcolor{mygreen1}{\scriptsize~+~Full Dataset(DAPO)@500} & 78.64 & 53.27 & 42.73 & 29.38 & 20.05 & 0.0 & 65.95 & 23.04\\
\rowcolor{mygreen2}{\scriptsize~+~Full Dataset(DAPO)@1k} & 78.93 & 53.26 & 41.88 & 30.21 & 19.85 & 0.0 & 66.09 & 22.99\\
\rowcolor{mygreen1}{\scriptsize~+~Learnability(GRPO)@500} & 77.57 & 59.05 & 50.19 & 30.63 & 27.55 & 5.00 & 68.31 & \textbf{28.34} \\
\rowcolor{mygreen2}{\scriptsize~+~Learnability(GRPO)@1k} & 79.12 & 59.03 & 47.19 & 29.17 & 26.96 & 2.78 & 69.07 & 26.52 \\
\rowcolor{mygreen1}{\scriptsize~+~Pass Rate(GRPO)@500} & 78.82 & 57.87 & 49.09 & 26.88 & 26.08 & 3.33 & 68.35 & 26.34 \\
\rowcolor{mygreen2}{\scriptsize~+~Pass Rate(GRPO)@1k} & 80.19 & 58.33 & 46.86 & 29.17 & 27.12 & 8.33 & 69.26 & 27.87 \\
\rowcolor{mygreen1}{\scriptsize~+~Influence(GRPO)@500} & 78.31 & 58.80 & 49.74 & 31.25 & 25.49 & 5.83 & 68.56 & {28.08} \\
\rowcolor{mygreen2}{\scriptsize~+~Influence(GRPO)@1k} & 78.58 & 58.82 & 47.73 & 23.96 & 25.08 & 8.33 & 68.70 & 26.28 \\
\rowcolor{mygreen1}{\scriptsize~+~\textbf{CROPI (Ours)}@500} & 80.55 & 58.43 & 48.12 & 26.25 & 26.86 & 4.17 & \textbf{69.49} & 26.35 \\
\rowcolor{mygreen2}{\scriptsize~+~\textbf{CROPI (Ours)}@1k} & 81.36 & 59.17 & 46.54 & 34.38 & 27.78 & 9.72 & \textbf{70.26} & \textbf{29.60} \\

\midrule

\rowcolor{verylightgreen1}{Qwen2.5-7B-Instruct} & 90.51 & 75.08  & 62.64 & 46.88 & 41.91 & 8.33 & 53.96 & 54.76 \\
\midrule
\rowcolor{mygreen1}{\scriptsize~+~Full Dataset(GRPO)@300} & 92.59 & 76.61 & 60.84 & 55.00 & 43.58 & 16.67 & 57.36 & 57.92 \\
\rowcolor{mygreen2}{\scriptsize~+~Full Dataset(GRPO)@600} & 93.54 & 77.39 & 60.97 & 52.50 & 45.49 & 13.33 & 57.44 & 56.74 \\
\rowcolor{mygreen1}{\scriptsize~+~\textbf{CROPI (Ours)}@300} & 92.13 & 77.92 & 65.39 & 58.75 & 45.64 & 14.17 & \textbf{57.46} & \textbf{62.07} \\
\rowcolor{mygreen2}{\scriptsize~+~\textbf{CROPI (Ours)}@600} & 92.89 & 78.35 & 63.70 & 55.62 & 45.78 & 17.50 & \textbf{58.63} & \textbf{59.66} \\

\midrule

\rowcolor{verylightgreen1}{R1-Distill-Qwen-1.5B} &77.73 & 72.83 & 60.07 & 53.13 & 41.67 & 20.83 & 43.93 & 75.28 \\
\midrule
\rowcolor{mygreen1}{\scriptsize~+~Full Dataset(GRPO)@150} & 77.42 & 73.64 & 63.77 & 53.75 & 41.62 & 17.50 & 44.16 & 75.53 \\
\rowcolor{mygreen2}{\scriptsize~+~Full Dataset(GRPO)@300} & 79.46 & 76.63 & 65.15 & 55.21 & 44.53 & 23.61 & 47.12 & \textbf{78.05} \\
\rowcolor{mygreen1}{\scriptsize~+~\textbf{CROPI (Ours)}@150} & 77.97 & 75.54 & 63.44 & 51.25 & 43.97 & 22.50 & \textbf{45.29} & \textbf{76.76} \\
\rowcolor{mygreen2}{\scriptsize~+~\textbf{CROPI (Ours)}@300} & 79.18 & 76.67 & 66.88 & 60.42 & 43.63 & 20.83 & \textbf{47.94 }& 77.92 \\

\bottomrule
\end{tabular*}
\end{table*}

\subsection{Setup}
\label{sec:exp_set}

We evaluate CROPI on mathematical reasoning tasks using several open-source, instruction-aligned models spanning different model scales and context sizes: Qwen2.5-1.5B-Instruct, Qwen2.5-7B-Instruct \citep{qwen2025qwen25technicalreport}, and Deepseek-R1-Distill-Qwen-1.5B \citep{guo2025deepseek}.
For simplicity, these models are referred to 1.5B, 7B, and 1.5B-R1, respectively.
Our training data consists of 47K unique problems aggregated from GSM8K-Train \citep{cobbe2021training}, MATH-Train\citep{hendrycks2021measuring}, and DeepScaleR-Preview-Dataset \citep{deepscaler2025}. We assess performance on a suite of standard benchmarks, including GSM8K-Test, MATH-Test, Gaokao2023EN \citep{mario_gaokao2023_math_en}, OlympiadBench \citep{he2024olympiadbench}, AMC23 \citep{mathai_amc23} and AIME24 \citep{mathai_aime24}. 
For influence-based data selection, a small validation set (max 100 examples) is sampled from a subset of these test sets.
We refer to tasks in which this validation set is employed for data selection in CROPI as "Targeted," while other test sets are designated as untargeted tasks ("Untar.").

We compared CROPI with several data selection baselines in selection ratio $\alpha=0.1$: Learnability \citep{bae2025online}, Pass Rate \citep{yu2025dapo}, Influence Function \citep{pruthi2020estimating} for global-level data selection, and the DAPO \citep{yu2025dapo} for batch-level data selection. 
Due to computational constraints, baseline comparisons are conducted on the 1.5B model. Full details regarding model versions, dataset construction, and hyperparameters are available in Appendix \ref{sec:appendix_exp_details}.
Unless otherwise stated, all reported \texttt{Acc.@step} numbers and training curves use a moving average with a sliding window of size $5$ to reduce the short-term variance of RL training.

\subsection{Main Results}
\label{sec:exp_result}

\begin{figure}[ht]
    \centering
    \includegraphics[width=0.50\textwidth]{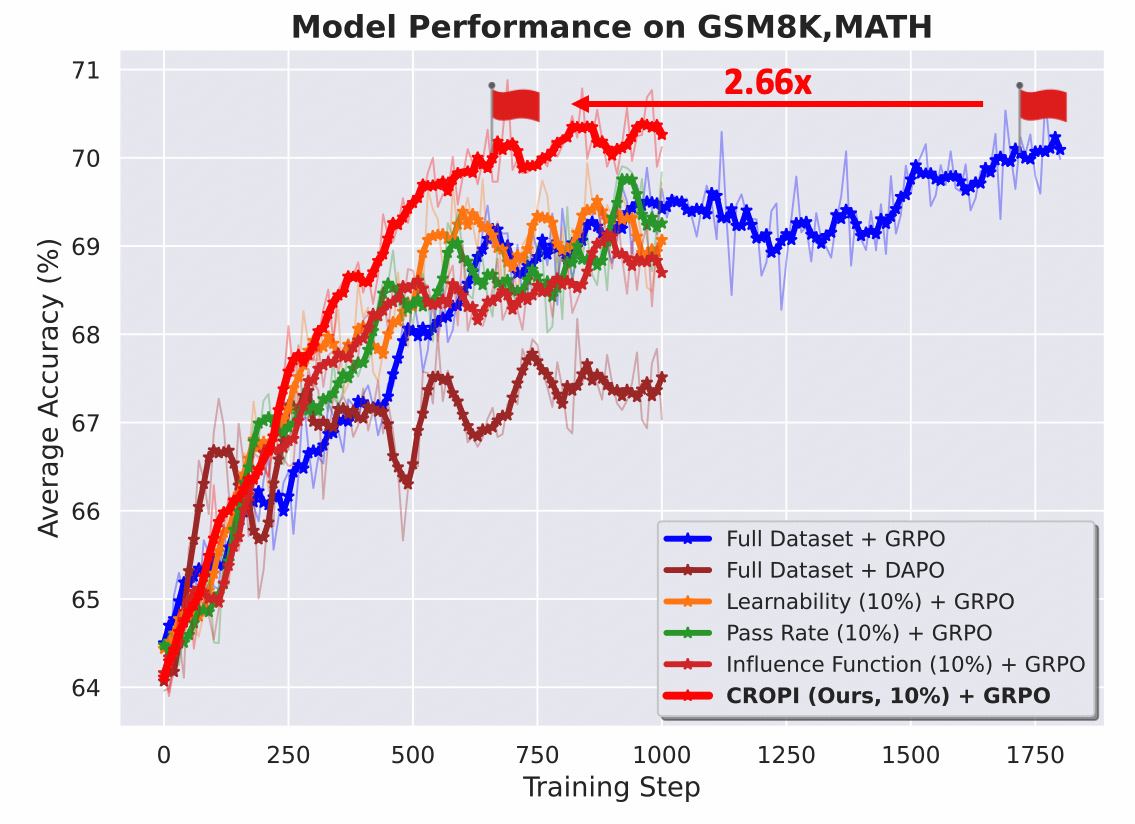}
    \caption{Training curves on the 1.5B setting. Curves are smoothed with a moving average of window size $5$. CROPI surpasses all other baselines and achieves a significant step-level acceleration ratio of $\bm{2.66\times}$ compared to full-data training while using only \textbf{10\%} of the data in each phase. 
    }
    \label{fig:1.5b_curve}
\end{figure}

As shown in Table \ref{tab:main_result}, CROPI consistently outperforms both full-data training and all baselines across different model scales in targeted tasks, demonstrating superior sample and step efficiency.
The improvements are particularly pronounced in the early stages of training.
For the 1.5B model, CROPI achieves a remarkable $\bm{2.66\times}$ \textbf{step-level speedup} compared to training on the full dataset on targeted tasks, as illustrated in Figure \ref{fig:1.5b_curve}. Here, using only 10\% of the training data in each phase means that at every phase we re-score the full training pool with the current policy, select the top 10\% prompts, and run GRPO only on that refreshed subset.
In contrast, while other data selection baselines show some initial gains, their performance plateaus quickly because they rely on a static estimation of data utility and fail to adapt to the evolving policy.
Meanwhile, CROPI also exhibits strong generalization capabilities.
We observe significant performance gains on "Untargeted" benchmarks---those not used for creating the validation set---indicating that the data selected by CROPI benefits the model's overall reasoning ability, not just performance on the targeted tasks.
Additional analyses in Appendix \ref{sec:appendix_convergence} and Appendix \ref{sec:appendix_random_baseline} further show that CROPI reaches its best smoothed accuracy earlier than standard GRPO and that replacing influence-based selection with random phase-wise selection leads to a clear performance drop.

As can be seen in Table \ref{tab:runtime}, due to the requirement to compute gradients over the entire dataset (after filtering out samples that are completely correct or incorrect), the data selection time for CROPI remains non-negligible even without new rollouts.
However, even after accounting for the selection time, the slowdown factor is only about 0.81x, and CROPI still achieves a 2.16x speedup overall.
Moreover, there remains significant room for optimization in our gradient computation process, such as selecting partial subsets, accelerating parallelization, or training a proxy scorer. 
Thus, the time speedup of CROPI could be further improved, highlighting its considerable potential.
For the 1.5B model, a direct wall-clock analysis shows that CROPI reaches 70\% targeted accuracy with a 2.17x speedup over standard GRPO; we report the corresponding convergence and fixed-time comparisons in Appendix \ref{sec:appendix_convergence}.

\begin{table}[ht]
    \centering
    \small
    \begin{tabular}{p{1.8cm}p{1.8cm}p{2.6cm}}
    \toprule
    \textbf{Time Cost} & \textbf{Select} & \textbf{Train}\\
    \midrule
    1.5B & 1.2h (19k) & 5.2h (200 steps) \\
    7B & 2.6h (18k) & 9.6h (200 steps) \\
    1.5B-R1 & 3.4h (17k) & 13.7h (100 steps) \\
    \bottomrule
    \end{tabular}
    \caption{Time costs of data selection and training for CROPI in each phase. We denote the number of prompts to process (gradient computation, projection, cosine similarity) in data selection and the optimization steps in the training stage in brackets. Time cost is measured on an 8-GPU (NVIDIA H100) machine. }
    \label{tab:runtime}
\end{table}

\vspace{-1em}
\section{Analysis}
\label{sec:ana}

This section provides an in-depth analysis of two key computational components of the CROPI framework: random projection and data selection.
We also provide an empirical analysis of off-policy gradient estimation in POPI in Appendix \ref{sec:off_policy_estimator}.

\subsection{Analysis on Sparse Random Projection}
\label{sec:ana_random_projection}

\begin{figure}[ht]
    \centering
    \includegraphics[width=0.50\textwidth]{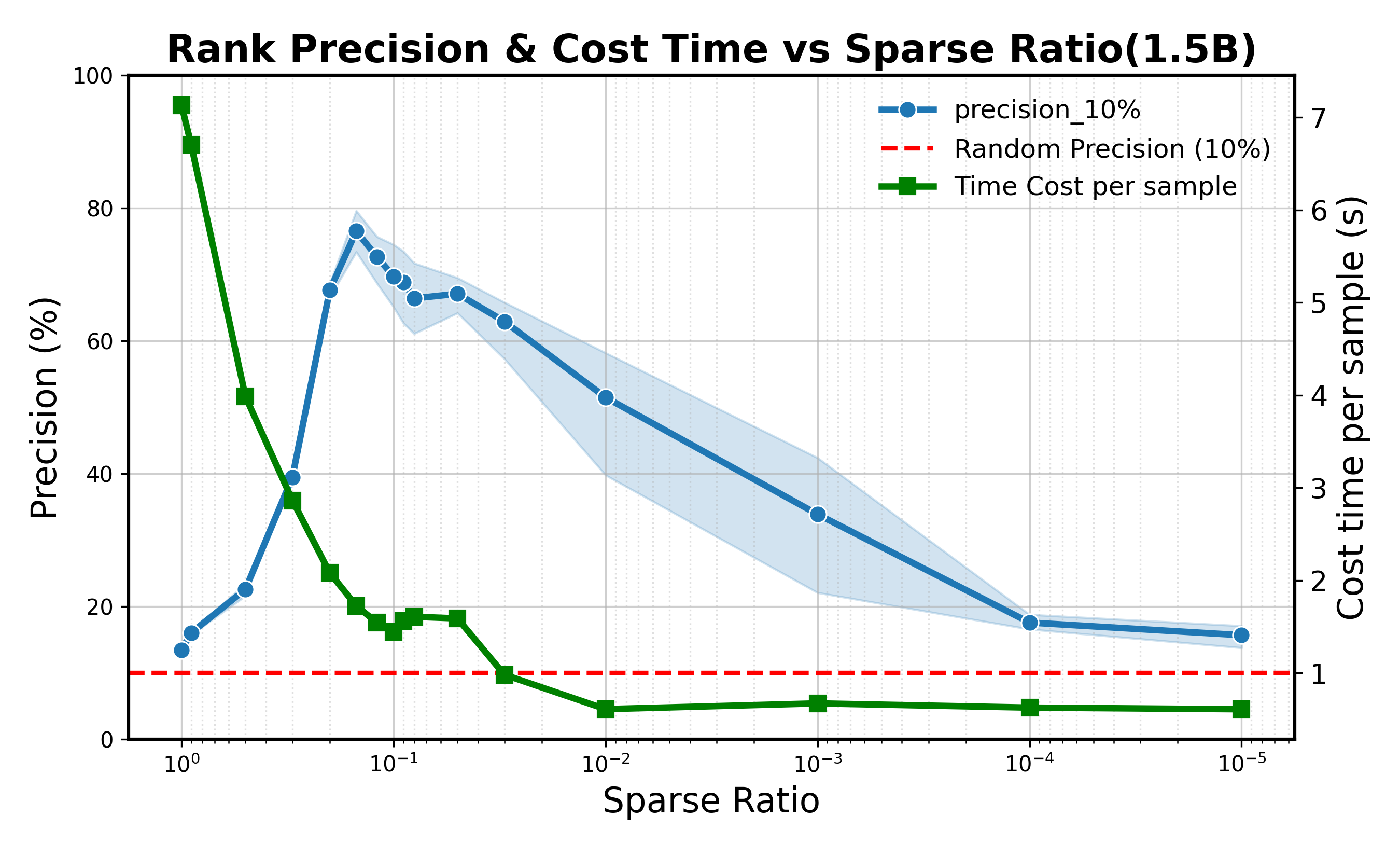}
    \caption{Rank preservation experiments for Sparse Random Projection. }
    \label{fig:proj_rank_preservation}
\end{figure}
As mentioned on Section \ref{sec:opi}, we use sparse random projection for the projection of full-parameter gradients.
Specifically, we randomly select a proportion of dimension of the gradient to perform random projection to avoid computation over the entire high-dimensional gradient. 
We define this proportion as sparse ratio.
However, through our experiments, we found that this dropout actually improves the preservation of inner products between gradient features after random projection.

We sampled 50 prompts from the training set and computed the GRPO gradients for the 1.5B model. 
We then selected a subset of gradient dimensions according to a predefined sparse ratio and applied random projection to these selected dimensions. 
For all gradients, we compute the pairwise cosine similarities before and after sparse random projection.
We define \texttt{precision@10\%} as the probability that, for each projected gradient, the top 10\% most similar gradients (based on cosine similarity) contain the top 10\% most similar gradients as measured by the full-parameter cosine similarities. This metric reflects the degree to which the random projection preserves the ranking of similarities among gradient features. A higher \texttt{precision@10\%} indicates better preservation.

Figure \ref{fig:proj_rank_preservation} presents the experimental results under the 1.5B Setting. We observe that when we directly apply random projection to the full gradient (i.e. sparse ratio equals 1) , the \texttt{precision@10\%} of similarity ranking is only around 13\%, comparable to random selection. However, when the sparse ratio is 0.1, the \texttt{precision@10\%} is significantly higher, reaching nearly 80\%.
This is a counter-intuitive phenomenon: under sparse random projection, the ranking preservation is actually better with less information. We hypothesize that this may be related to the presence of numerical noise in the gradients: random projection can amplify such noise. 
\textbf{\textit{Sparsity, while masking some information, filters out much of the numerical noise and achieves a better signal-to-noise ratio around a sparse ratio of 0.1}}.
We include more analysis in Appendix \ref{sec:add_ana_proj}.

\subsection{Analysis on data selected by CROPI}
\label{sec:ana_data}

We analyze the training data selected by CROPI under the 1.5B setting. 
Specifically, we examine the top-100 and bottom-100 training samples ranked by POPI scores, given validation sets GSM8K and MATH. 
For each checkpoint (training steps: 0, 200, 400, 600, 800), we evaluate both the semantic similarity between selected samples and the validation set, as well as the model's pass rate on these samples.

\begin{figure}[t]
    \centering
    \includegraphics[width=0.95\columnwidth]{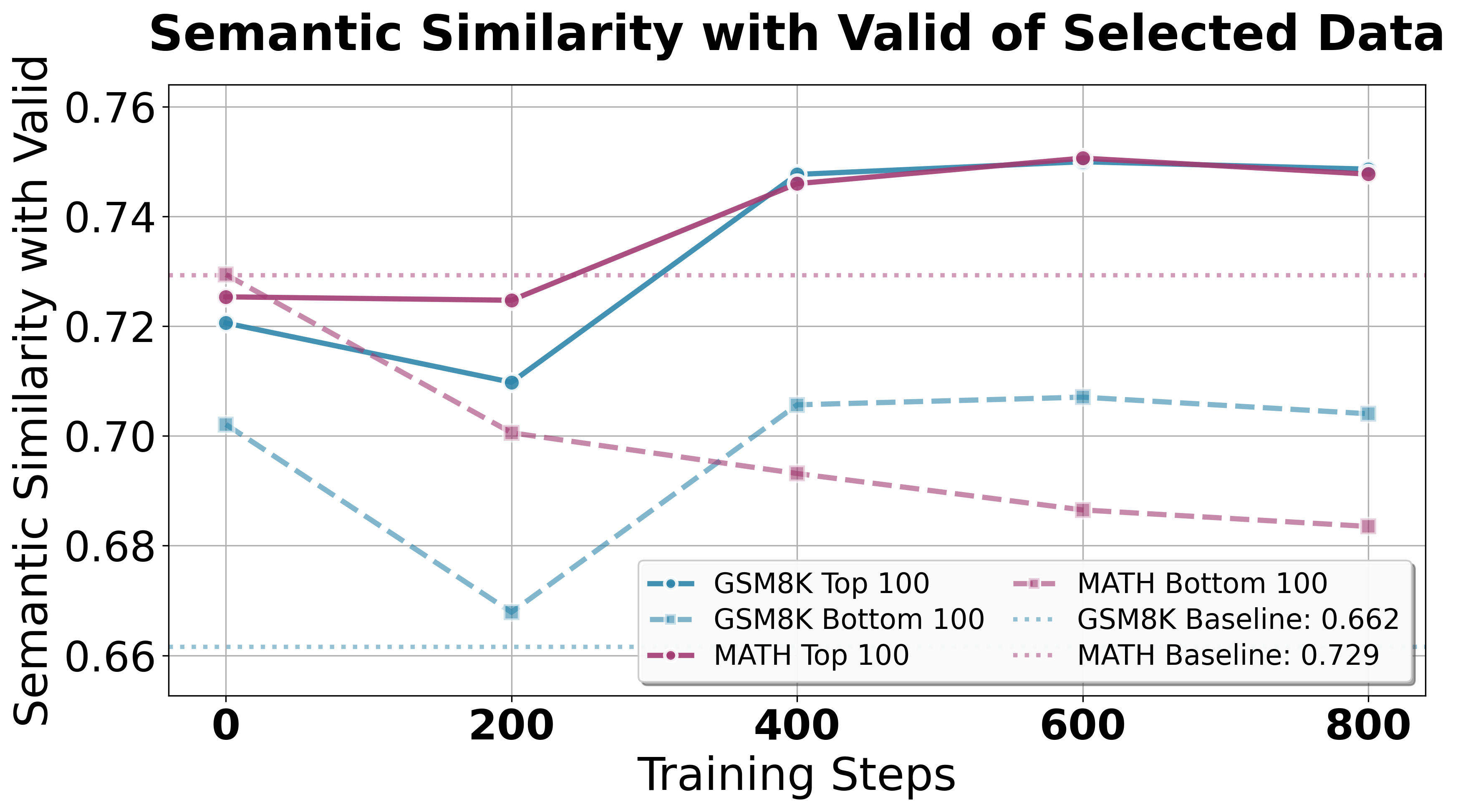}
    \caption{Semantic similarity between top-100 and bottom-100 training prompts selected by POPI and the validation set.}
    \label{fig:cropi_sel_semantic_valid}
\end{figure}

\noindent \textbf{Sematic Correlation.} 
For semantic analysis, we compute embeddings for each prompt using \texttt{BGE-large-en-V1.5} \citep{bge_embedding}. The average embedding of the validation set is compared to the embeddings of top-100 and bottom-100 samples via cosine similarity. 
As shown in Figure~\ref{fig:cropi_sel_semantic_valid}, data selected by POPI exhibits higher semantic similarity to the validation set compared to both randomly sampled (baseline) and bottom-100 samples.
These findings suggest that POPI leverages \textbf{\textit{latent relationships between gradient and semantic spaces}} to automatically identify training samples most relevant to the validation set.

\begin{figure}[t]
    \centering
    \includegraphics[width=\columnwidth]{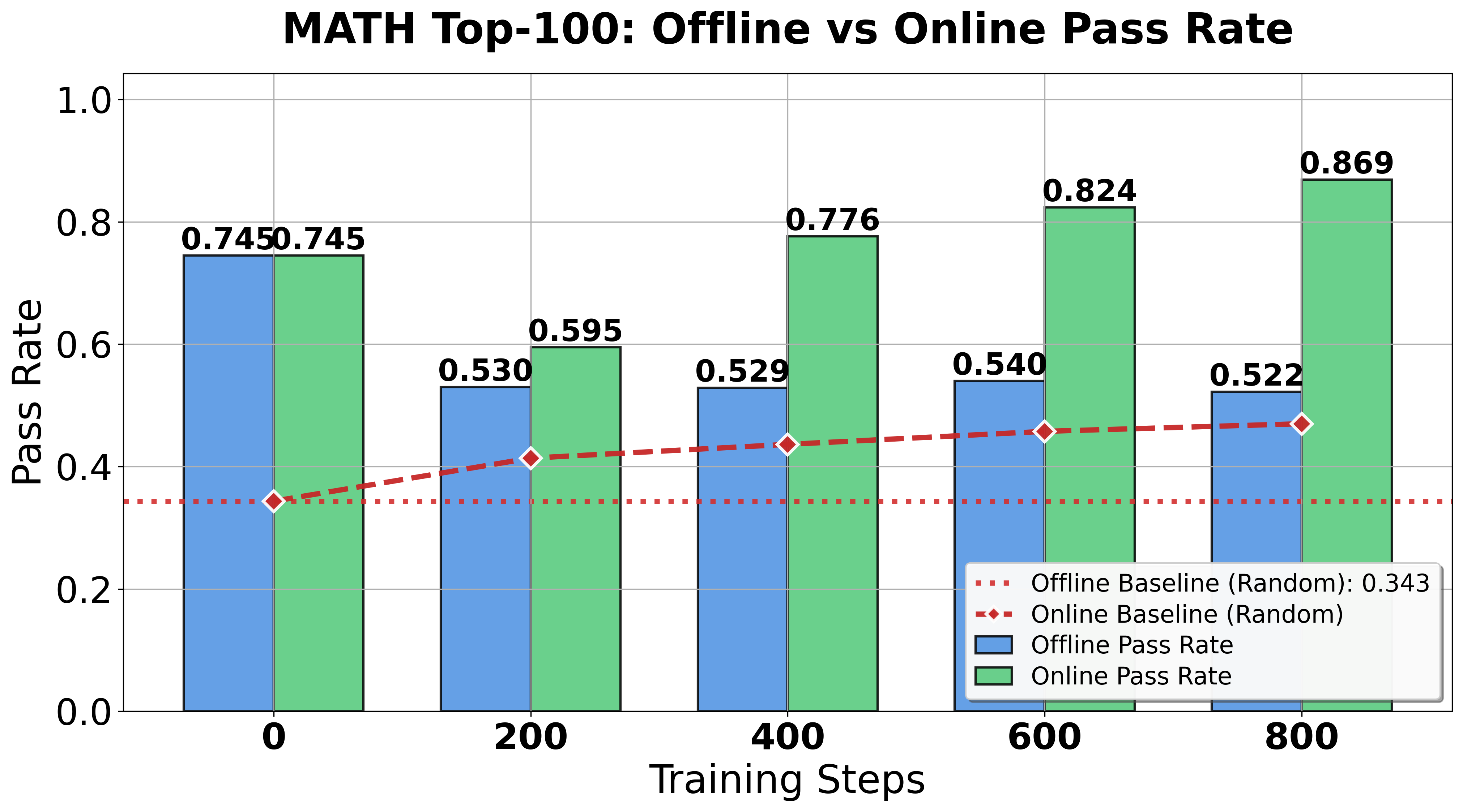}
    \caption{MATH Top-100: Offline vs Online Pass Rate. This figure compares the pass rates of the top-100 prompts selected using the MATH validation set. The \textbf{Offline Pass Rate (blue)} shows the performance of the base model on these prompts, indicating their inherent difficulty. The \textbf{Online Pass Rate (green)} shows the performance of the model at the current training step, demonstrating its learning progress on the curated data.}
    \label{fig:cropi_sel_pass_rate_combined}
\end{figure}

\noindent \textbf{Pass Rate. } 
To analyze the pass rate of the selected data, we contrast the performance of the top-100 prompts (selected using the MATH validation set) on the base model versus the evolving online model. As illustrated in Figure~\ref{fig:cropi_sel_pass_rate_combined}, we plot both the offline pass rate (pass rate on the base model) and the online pass rate (pass rate on the current training checkpoint). 
The offline pass rate (blue bars) shows a downward trend after the initial step, dropping from 0.75 to around 0.53. This indicates that as training progresses, CROPI selects problems that are increasingly difficult for the original base model.
In stark contrast, the online pass rate (green bars) exhibits a strong upward trend, rising from 0.75 to 0.87. This demonstrates that \textbf{\textit{ while the selected problems are challenging, they fall within the current model's learning frontier}}. 
The model effectively learns to solve them, reflecting CROPI's ability to dynamically select data that maximizes learning efficiency. 
Ultimately, the model is trained on data with a high online pass rate (in the 0.6 to 0.9 range), which corresponds to the difficulty interval where performance improvement is most pronounced.

In Appendix~\ref{sec:more_select}, we provide a detailed breakdown of data sources, knowledge categories, and the diversity of selected data.
These results demonstrate that CROPI not only achieves strong performance but also offers a degree of interpretability.


\section{Conclusion}


This paper introduced a principled data selection method for RLVR using influence functions. To circumvent the prohibitive cost of online rollouts for LLMs, we developed a novel off-policy estimation technique that evaluates data influence using offline trajectories. 
Our curriculum learning framework, CROPI, leverages this method to select the most impactful data for training. 
Experiments demonstrate that CROPI significantly accelerates learning, achieving a remarkable speedup on RL training while using substantially less data than full-dataset training. 
Our work validates that influence-based data selection is a theoretically-grounded and highly efficient alternative to common heuristics, paving the way for more scalable and effective training of large reasoning models.

\section*{Limitations}

\noindent \textbf{Theoretical Limitations.}
In this paper, we restrict our analysis to first-order influence estimation under the SGD optimizer. 
Future work could extend this framework to encompass a broader range of optimizers and influence estimation approaches. 
While we employ off-policy gradient estimates to compute influence, we do not provide a theoretical analysis of the associated errors, bias, or variance; addressing these aspects is left for future research. 

\noindent \textbf{Limitations of Offline Trajectories.}
In this study, we estimate the gradients solely using trajectories from the base model. 
This choice results in certain training prompts having zero gradient -- specifically, those for which the base model’s predictions are entirely correct or incorrect (for these cases, the GRPO advantage is zero), and thus they lack gradient information during online gradient computation. Further investigations could consider incorporating positive (strong) or negative examples to ensure that all prompts possess non-zero gradient signals.
 Additionally, we do not explore the use of rollouts generated by smaller models to estimate the gradients of larger models, nor do we investigate reusing rollouts collected during the training process (e.g., via a replay buffer). These directions remain open for future work.

\noindent \textbf{Limitations of the Experimental Setting.}
In terms of task scope, our experiments are limited to single-turn mathematics question answering scenarios; extension to multi-turn dialogues and other types of reasoning, agentic, or multi-modal tasks is a promising direction for future studies.
In terms of scale, our empirical validation predominantly focuses on models of 1.5B and 7B parameters, with training steps limited to fewer than 1000. 
Extension to larger-scale models and longer training durations is left to future work.

\noindent \textbf{Numerical error in gradient computation.} 
As we utilize \texttt{float16} format to conduct gradient-related calculations, numerical error cannot be ignored and might be amplified by random projection. 
We introduce a random dropout operation before random projection as a way to obtain better rank preservation. 
Additional details can be found in Appendix \ref{sec:add_ana_proj}.




\newpage
\bibliography{main}

\appendix

\onecolumn

\section{Derivation of First-order Influence Function}
\label{sec:inf_theory}

In this Section, we will restate the first-order influence estimation in \texttt{TracIn} \cite{pruthi2020estimating} in RL context.

In RL context, our goal is to maximize performance function $J(\theta)=\mathbb{E}_{s_0\sim \rho_0(\cdot), \tau \sim \pii }[R(\tau)] $ \citep{sutton1999reinforcement}.
In LLM, the $s_0$ corresponds to the input query.
We rewrite the RL objective under a test query $s_0'$:

\begin{align}
    & \max_\theta~\mathbb{E}_{s_0'\sim q(\cdot)} [J(\theta; s_0')],\\
    & \text{~where~} J(\theta; s_0') \coloneqq \mathbb E_{\tau' \sim \pi_\theta(\cdot|s_0')} [R(\tau')]
\end{align}

Consider a policy $\pi_{\theta^l}$ at optimization step $l$. 
Using first-order Taylor Expansion, the objective function in next iteration step $J(\theta^{l+1};s_0')$ can be rewritten as: 

\begin{align}
    J(\theta^{l+1}; s_0') &= J(\theta^{l};s_0') +  \langle \nabla_\theta J(\theta^l; s_0'),~~\theta^{l+1} - \theta^{l} \rangle  + \ O(||\theta^{l+1} - \theta^{l}||^2 ) \\
    &\approx J(\theta^{l};s_0') +  \langle \nabla_\theta J(\theta^l; s_0'),~~\theta^{l+1} - \theta^{l} \rangle 
\end{align}

Consider training the policy $\pi_{\theta^l}$ using Stochastic Gradient Ascent (SGA) with batch size 1 and learning rate $\eta_l$.  
At iteration $l$, let $s_0$ be the initial state sampled as the training data point.  
Since the objective in reinforcement learning is to maximize the performance function $J$, the update uses a positive sign before the policy gradient:

\begin{align}
\theta^{l+1} &= \theta^{l} + \eta_t \nabla_\theta J (\theta^l; s_0)\\
    J(\theta^{l+1}; s_0') &\approx J(\theta^{l};s_0')  + \eta_t \langle \nabla_\theta J(\theta^l; s_0'),~~\nabla_\theta J (\theta^l; s_0) \rangle 
\end{align}

Consider training a policy with $N$ training prompts, denoted by $\{s_0^{(i)}\}_{i=1}^{N}$. 
Following the approximation of $\text{TracInCP}$ as proposed in \citep{pruthi2020estimating}, we treat one epoch of training as a single optimization step with a large batch size $N$, and, for simplicity, we ignore parameter updates within the epoch by assuming the parameters remain constant throughout. 
Let the parameters at the start and end of the $m$-th epoch be represented as $\theta^{(m)} = \theta^{l_m}$ and $\theta^{(m+1)} = \theta^{l_{m+1}}$, respectively, where the actual number of update steps within the epoch is $l_{m+1} - l_m$.

Under this approximation, the parameter update after one epoch can be expressed as
\begin{align}
    \theta^{(m+1)} &\approx \theta^{(m)} + \eta_t \sum_{i=1}^{N} \nabla_\theta J(\theta^{(m)}; s_0^{(i)}),
\end{align}
and for a test sample $s_0'$, the change in the objective function is approximated by
\begin{align}
    J(\theta^{(m+1)}; s_0') &\approx J(\theta^{(m)}; s_0') + \eta_t \sum_{i=1}^{N} \langle \nabla_\theta J(\theta^{(m)}; s_0'),~\nabla_\theta J(\theta^{(m)}; s_0^{(i)}) \rangle.
\end{align}

We define the first-order influence of a training prompt $s_0^{(i)}$ on the model checkpoint $\pi_{\theta^{(m)}}$ with respect to a test sample $s_0'$ as:
\begin{align}
    \text{Inf}(\pi_{\theta^{(m)}}; s_0^{(i)}, s_0') \coloneqq \langle \nabla_\theta J(\theta^{(m)}; s_0'),~\nabla_\theta J(\theta^{(m)}; s_0^{(i)}) \rangle.
\end{align}

If we have multiple test samples, denote the distribution of test query as $q(\cdot), s_0'\sim q(\cdot)$, then the objective function becomes $J(\theta;q)=\mathbb{E}_{s_0'\sim q(\cdot), \tau '\sim \pii(\cdot|s_0') }[R(\tau')] $.
With similar derivation, we can rewrite the influence of  training prompt $s_0^{(i)}$ with respect to test distribution $q$ as:
\begin{align}
    \text{Inf}(\pi_{\theta^{(m)}}; s_0^{(i)}, q) \coloneqq \langle \nabla_\theta J(\theta^{(m)}; q),~\nabla_\theta J(\theta^{(m)}; s_0^{(i)}) \rangle.
\end{align}

In practical implementations, to prevent data leakage, we allocate a small subset of the training data as a validation set and use the validation samples to measure the influence of the training data.

\section{Derivation of Off-Policy Gradient}
\label{sec:opi_theory}

In order to estimate the online influence with offline trajectories, we need to first approximate the policy gradient using offline trajectories, this is equivalent to the off-policy policy gradient estimation. 

We first write the vanilla policy gradient in on-policy form \citep{sutton1999reinforcement}: 

\begin{align}
    \nabla_\theta J(\theta) &= \mathbb E_{\tau \sim \pii}[\sum_{t=0}^{|\tau|-1}A_t \nabla_\theta \log \pii (x_t | s_t)]
\end{align}

Since we are using initial trajectories $\tau \sim \beta = \pi_{\theta_0} $ to estimate the gradient, using importance sampling: 

\begin{align}
    \nonumber \nabla_\theta J(\theta;s_0) &= \mathbb E_{\tau \sim \pii(\cdot|s_0) }[ \sum_{t=0}^{|\tau|-1}A_t \nabla_\theta \log \pii (x_t | s_t)]
    \\
    &= \mathbb E_{\tau \sim \beta(\cdot|s_0) }[ \rho_\theta(\tau | s_0) \sum_{t=0}^{|\tau|-1}A_t \nabla_\theta \log \pii (x_t | s_t)] \\
    \text{where}~~\rho_\theta(\tau|s_0) &= \frac{ \pii(\tau|s_0) }{\beta(\tau|s_0)}
\end{align}

However, for multi-step trajectory generation, the variance of the importance weight $\rho_\theta(\tau|s_0)$ can become extremely large.
This high variance arises due to the product of many likelihood ratios over the steps, leading to instability and poor sample efficiency in the gradient estimate.
We therefore adopt a more widely used off-policy gradient estimator from \citep{schulman2015trust, schulman2017proximal} that uses importance sampling at the token level:

\begin{align}
    \grad_{\theta} J (\theta;s_0) 
   \nonumber &=\nabla_\theta J(\pi_\theta; s_0) \\
   \nonumber &= \nabla_\theta [ J(\beta; s_0) + \EE_{s\sim 
   \nonumber {d^{\pii}(\cdot|s_0)}, x\sim \pii(\cdot|s)} [A^{\beta }(s,x)] )]\\
   \nonumber &\approx \nabla_\theta [ J(\beta; s_0) + \EE_{s\sim 
   \nonumber \textcolor{red}{d^\beta(\cdot|s_0)}, x\sim \pii(\cdot|s)} [A^{\beta }(s,x)] )]\\
   \nonumber &= \nabla_\theta [ J(\beta; s_0) + \EE_{s\sim d^\beta(\cdot|s_0), x\sim \beta(\cdot|s)} [ \frac{\pii(x|s) }{\beta(x|s)}A^{\beta }(s,x)] ] \\
   \nonumber &= \nabla_\theta \EE_{s\sim d^\beta(\cdot|s_0), x\sim \beta(\cdot|s)} [ \frac{\pii(x|s) }{\beta(x|s)}A^{\beta }(s,x)] ) \\
   \nonumber &= \EE_{s\sim d^\beta(\cdot|s_0), x\sim \beta(\cdot|s)} [ \frac{\pii(x|s) }{\beta(x|s)} A^{\beta }(s,x) \nabla_\theta \log \pii(x|s)] ) \\
   \nonumber &= \EE_{t, s_t \sim d^\beta(s_t|s_0), x_{t}\sim \beta(x_t|s_t)} \left[ \frac{\pii(x_t|s_t)} {\beta(x_t|s_t)}  A^{\beta}_t    \grad_\theta \log \pii(x_t|s_t) \right] \\
   \nonumber &= \EE_{t, s_t \sim d^\beta(s_t|s_0), x_{t}\sim \beta(x_t|s_t)} \left[ \rho_t^{\pii}  A^{\beta}_t   \grad_\theta \log \pii(x_t|s_t) \right]~~~\left( \rho^{\pii}_t \coloneqq \frac{\pii(x_t|s_t)}{\beta(x_t|s_t)} \right)\\
   \nonumber &=\EE_{\tau \sim \beta(|s_0)} \left[ \frac{1}{|\tau|} \sum_{t=0}^{|\tau|-1} \rho_t^{\pii} A^{\beta}_t  \grad_\theta \log \pii(x_t|s_t) \right] \\
   \nonumber &= \EE_{\{\tau_k\}_{k=1}^{K} \sim \beta(|s_0)}\left[ \frac{1}{K} \sum_{k=1}^{K}  \frac{1}{|\tau_k|} \sum_{t=0}^{|\tau_k|-1} \rho_{k,t}^{\pii}A^{\beta}_{k,t} \grad_\theta \log \pii(x_{k,t}|s_{k,t})  \right] \\
    &= \EE_{\{\tau_k\}_{k=1}^{K} \sim \beta(|s_0)}\left[ \frac{1}{K} \sum_{k=1}^{K}  \frac{1}{|\tau_k|} \sum_{t=0}^{|\tau_k|-1} \nabla_\theta \rho_{k,t}^{\pii}A^{\beta}_{k,t}   \right]
\end{align}

Thus, our off-policy gradient estimator can be formulated as:

\begin{align}
    \widehat{g}_\beta (\theta, s_0, \{\tau_k\}_{k=1}^K) \approx &  \frac{1}{K} \sum_{k=1}^{K}  \frac{1}{|\tau_k|} \sum_{t=0}^{|\tau_k|-1} \nabla_\theta \rho_{k,t}^{\pii}\widehat{A}^{\beta}_{k,t}  \\
    \text{where~~} 
\end{align}

For group normalized advantage estimator in GRPO, we have: 

\begin{align}
    \rho^{\pii}_{k,t} &= \frac{\pii(x_{k,t}|s_{k,t})}{\beta(x_{k,t}|s_{k,t})} \\
    \widehat{A}_{k,t}^{\beta} &= \frac{ R(\tau_k) - \widehat{\EE}_{\beta} [  R(\tau) ]} {\widehat{\sigma}_{\beta}[R(\tau)]}
\end{align}

$\widehat{\EE}_{\beta}$ denotes empirical mean under policy $\beta$, i.e. $\widehat{\EE}_{\beta} [  R(\tau) ]=\frac{1}{K} \sum_{k=1}^{K}[R(\tau_k)], \tau_k \sim \beta(\cdot|s_0), k=1,2,...,K$.
$\widehat{\sigma}_{\beta}$ denotes the empirical standard value under policy $\beta$: 
$\widehat{\sigma}_{\beta}[R(\tau)]=\sqrt{ \widehat{\EE}_{\beta} \left[  (R(\tau) - \widehat{\EE}_{\beta} [  R(\tau) ] )^2 \right] }$. 

The approximations employed here generally require that $\pii$ and $\beta$ are relatively close. 
In TRPO \cite{schulman2015trust}, $\beta=\piiold$ is constrained to be close to $\pii$ in terms of the Kullback-Leibler (KL) divergence.  
In the context of LLM RLVR, when performing GRPO \cite{shao2024deepseekmath} training, a KL loss is introduced to constrain the divergence between the training policy and the initial policy.  
Experimental results demonstrate that the estimated KL values during RLVR training are usually maintained below 0.1, which ensures the reasonableness of the approximation to a certain extent.

\twocolumn

\section{Related Works}
\label{sec:related}


\subsection{Data Selection for RLVR}
Recently, how to automatically select high-quality data for RLVR has become a research topic of increasing interest.
According to the time granularity of data selection, existing methods can be roughly divided into three categories: global-level, batch-level and phase-level. 
The first is the global-level granularity, which selects data based on the initial policy over the entire dataset \cite{zhao2025ufo, wang2025reinforcement}. Although this approach is easy to implement, it cannot capture dynamic checkpoint information and often requires warm-up training. 
The second granularity is at the batch-level, where data selection is performed within each training batch. This method can better capture the dynamics of model training; however, due to the lack of long-term planning, it often introduces high variance that can lead to unstable training. 
Representative methods include DAPO \citep{yu2025dapo} and ODF \citep{bae2025online}.
The third category is a compromise, where data selection is performed every certain number of training steps, which we refer to as phase-level (or curriculum) data selection. 
This approach strikes a balance between capturing training dynamics and enabling long-term planning, but may still face challenges such as unstable training distributions. 
An example of this method is \cite{xi2024traininglargelanguagemodels}.

As for the utility estimator, previous works use various ways to measure the utility of each data point, mainly focusing on heuristics around the difficulty and uncertainty of the training query $s_0$. 
In order to select prompts with proper difficulties, some researchers use correctness rate of various trajectories as the signal of the difficulty \citep{li2025limr, yu2025dapo, bae2025online, sun2025improving}. 
As an example, \citet{bae2025online} suggests filtering data points with pass rate around $0.5$ to construct the batch for GRPO training.
\citet{zhao2025ufo} estimate prompt difficulty based on the model's confidence (likelihood), and select moderately difficult prompts for reinforcement learning (RL) training. 
Another heuristic involves selecting data according to the uncertainty arising from data perturbation. For instance, \citet{wang2025reinforcement} train with prompts whose associated historical trajectories exhibit the highest reward variance.

These methods generally need to obtain prompt trajectories to assess the utility of each prompt, but online rollouts with LLMs are costly in terms of computation and time.
In order to avoid online rollouts, these methods have made significant sacrifices in online estimation: either selecting data based on the initial policy, which is a global-level data selection, or using the pass rate signal from the data buffer as the basis for data filtering. 
Such compromises largely neglect the dynamics of RL training, for instance, the changes in the pass rate of the same data point across different stages of training.

\subsection{Influence Functions for LLM Data Selection}
Influence function \citep{koh2017understanding} is a gradient-based data attribution method derived from variantion analysis of objective function. 
Influence functions and their variants have been widely applied in the Pre-Training (PT) and Supervised Fine-Tuning (SFT) stages due to their strong theoretical guarantees and empirical effectiveness \cite{Engstrom2024dsdm, wang2024greats, gu2024data, grosse2023studying}. \citet{Engstrom2024dsdm} proposed the datamodel framework leveraging an efficient influence function estimation \cite{park2023trak}, which is inspired by influence function, to select pretraining data.
LESS \citep{xia2024less} is the first to introduce first-order influence functions into the post-training of large language models (LLMs), thereby improving the training efficiency of supervised fine-tuning (SFT). \citet{wang2024greats} further refine LESS by extending it to the batch-level; however, its application remains limited to the SFT stage.  
Nonetheless, leveraging influence function theory to guide data selection for RLVR remains a challenging problem. 
This is mainly due to the fact that, in RL settings, rewards and gradients for the data typically require rollouts to obtain, which is computationally expensive for LLMs.  
A concurrent work \citet{hu2025snapshotinfluencelocaldata} explores online data selection using influence functions in RLHF and demonstrates promising empirical results across multiple settings. However, their setting differs substantially from RLVR, and the target function used for influence estimation is defined for a different training objective.


\section{Algorithmic pseudocode of CROPI}
\label{sec:alg_cropi}

We formulate the process of our proposed method CROPI in Algorithem \ref{alg:cropi}. 

\begin{algorithm}[htbp]
\caption{Algorithmic pseudocode of CROPI}
\begin{algorithmic}
\label{alg:cropi}
\REQUIRE Training dataset $\mathcal{D}_{\text{tr}}$, Validation datasets $\{\mathcal D_{\text{val}, j}\}_{j=1}^{V}$, Base LLM $\pi_{\theta_0}$, Selection ratio $\alpha$.
number of phases $M$, training steps per phase $E$.
\ENSURE Output final policy $\pi_{\theta^{(M)}}$
\STATE Load initial policy $\pi_{\theta^{(0)}} \leftarrow \pi_{\theta_0}$
\FOR{$m=0,1,...,M-1$}
    \FOR {$s_0^{(i)}\in\mathcal{D}_{\text{tr}}, i=1,2,...,N$}
\STATE Compute POPI score~$U_{\text{POPI-R}}(\pi_{\theta^{(m)}}; s_0^{(i)})$;
    \ENDFOR
    \STATE Select training subset \\ $\mathcal{D}^{(m)} \leftarrow \arg\max\limits_{|\mathcal S|=\lfloor \alpha |\mathcal{D}_{\text{tr}}| \rfloor } \displaystyle\sum_{s_0 \in \mathcal S}U_{\text{POPI-R}}(\piit{m}; s_0)$ 
    \STATE Optimize the policy with GRPO \\
    $\piit{m+1} \leftarrow \text{GRPO}(\piit{m}; E)$
\ENDFOR
\RETURN $\pi_{\theta^{(M)}}$
\end{algorithmic}
\end{algorithm}

\section{Additional Experimental Details}
\label{sec:appendix_exp_details}

\subsection{Implementation Details}
\noindent \textbf{Base Models.} We use Qwen2.5-1.5B-Instruct, Qwen2.5-7B-Instruct \citep{qwen2025qwen25technicalreport}, and Deepseek-R1-Distill-Qwen-1.5B \citep{guo2025deepseek}, which we refer to as 1.5B, 7B, and 1.5B-R1, respectively. We use the official prompt template from Qwen-Math \citep{yang2024qwen2} for the Qwen2.x models and the DeepSeek-R1 template \citep{guo2025deepseek} for 1.5B-R1. For more details on prompts, please see Appendix \ref{sec:prompt_temp}.

 \noindent \textbf{Datasets.} Our primary training set is the intersection of GSM8K-Train \citep{cobbe2021training}, MATH-Train \citep{hendrycks2021measuring}, and DeepScaleR-Preview-Dataset \citep{deepscaler2025}, containing 47K unique mathematical queries. 
 For the 1.5B-R1 model experiments, we randomly sampled 25K queries from this set to reduce training time.

\noindent  \textbf{Validation Sets.} 
To enable influence-based selection, we create a validation set by allocating 20\% of each designated test set, with a maximum cap of 100 examples per set to prevent its use in training. 
To mitigate data leakage, all reported test results exclude these validation examples. 
The validation set composition varies by model:
\begin{itemize}
    \item \textbf{1.5B}: GSM8K-Test, MATH-Test
    \item \textbf{7B}: GSM8K-Test, MATH-Test, OlympiadBench, AIME24
    \item \textbf{1.5B-R1}: GAOKAO23EN, AMC23, OlympiadBench, AIME24
\end{itemize}
The remaining test sets are used to evaluate generalization performance. 
We divide the test benchmarks into two categories. 
Targeted tasks (or "Targeted") refer to those whose domains are represented in the validation set. 
All other test sets are designated as Untargeted tasks ("Untargeted"). To evaluate the in-domain and out-of-distribution (OOD) generalization of CROPI, we report the average accuracies on Targeted and Untargeted tasks, respectively, under various settings. 
Note that reported test results exclude performance on the validation sets.

\subsection{Hyperparameters of CROPI}
We use the \texttt{VeRL} \citep{sheng2024hybridflow} framework for training. 
For rollout engine, we use \texttt{vllm} \citep{kwon2023efficient}. 
For the POPI computation process, we set the sparse ratio of the random projection to 0.01. 
We use TRAK \citep{park2023trak} for efficient random projection on GPU. 
The data selection ratio is $\alpha=0.1 $. 
The number of update steps per training phase, $E$, is set to 200 for the 1.5B and 7B models and 100 for the 1.5B-R1 model. 
The total training steps for the 1.5B, 7B, and 1.5B-R1 models were 1000, 600, and 300, respectively.

\subsection{Baselines in Main Results}
We compare CROPI against the following baselines, with all hyperparameters aligned:
\begin{enumerate}
    \item \textbf{Learnability} \citep{bae2025online}: Utility is estimated as $U(\pii; s_0)=p(1-p) $, where $p$ is the offline pass rate of a prompt.
    \item \textbf{Pass Rate} \citep{muennighoff2025s1simpletesttimescaling, yu2025dapo}: Utility is defined as $U(\pii; s_0)=\mathbb{I}_{0<p<1} $, where $p$ is the offline pass rate of a prompt.
    \item \textbf{Influence Function} \citep{pruthi2020estimating}: Standard first-order influence function estimation. This baseline is equivalent to CROPI with only 1 phase. 
\end{enumerate}
For the above baselines, data selection is performed once globally using the initial policy $\pi_{\theta_0}$ at a selection ratio of $\alpha=0.1 $. We also include \textbf{DAPO} \citep{yu2025dapo}, a batch-level filtering method that removes samples with perfect or zero pass rates and replaces them from a buffer of historical trajectories.
All reported \texttt{Acc.@step} metrics in the main text are computed from evaluation curves smoothed by a moving average with window size $5$.

\subsection{Additional Convergence and Time-based Comparison}
\label{sec:appendix_convergence}

To better characterize the final-stage behavior of CROPI, we extended the 1.5B experiment beyond the initial 1k-step budget and compared the best smoothed targeted accuracy achieved within 2000 training steps. The results in Table \ref{tab:peak_compare} show that CROPI reaches its peak performance earlier and attains a slightly higher best accuracy than standard GRPO within the same step budget.

\begin{table}[ht]
\centering
\small
\caption{Peak targeted accuracy within 2000 training steps on the 1.5B setting. Accuracy is computed from curves smoothed with a moving average of window size $5$.}
\label{tab:peak_compare}
\begin{tabular}{lcc}
\toprule
Method & Step & Peak Accuracy (\%) \\
\midrule
CROPI (Ours) & 1130 & 71.05 \\
Standard GRPO & 1790 & 70.17 \\
\bottomrule
\end{tabular}
\end{table}

We also evaluated both methods under fixed wall-clock budgets. As shown in Table \ref{tab:time_budget}, CROPI consistently outperforms standard GRPO at 12h, 18h, and 24h, which complements the step-level speedup reported in the main text.

\begin{table}[ht]
\centering
\small
\caption{Targeted accuracy (\%) under fixed wall-clock budgets for the 1.5B setting.}
\label{tab:time_budget}
\begin{tabular}{lccc}
\toprule
Method & 12h & 18h & 24h \\
\midrule
Standard GRPO & 67.73 & 68.84 & 69.34 \\
CROPI (Ours, 10\%) & 68.65 & 69.73 & 69.91 \\
\bottomrule
\end{tabular}
\end{table}

\subsection{Random Selection Baseline}
\label{sec:appendix_random_baseline}

To isolate the contribution of influence-based selection itself, we additionally evaluate a phase-wise random baseline that keeps the CROPI training schedule unchanged but replaces influence ranking with random selection of 10\% of the training data at each phase. Table \ref{tab:random_baseline} shows that random phase-wise selection underperforms both full-data GRPO and CROPI, indicating that the gain does not come from simply reducing the training set size.

\begin{table}[ht]
\centering
\small
\caption{Ablation with phase-wise random selection on the 1.5B setting.}
\label{tab:random_baseline}
\begin{tabular}{lcc}
\toprule
Method & Acc.@500 & Acc.@1000 \\
\midrule
Full Data + GRPO & 68.04 & 69.42 \\
CROPI w. Random Selection (10\%) & 67.29 & 68.04 \\
CROPI (Ours) & 69.49 & 70.26 \\
\bottomrule
\end{tabular}
\end{table}

\subsection{Prompt Templates for Evaluation \& RL Training}
\label{sec:prompt_temp}

For experiments on Qwen2.5-1.5B-Instruct (1.5B) and Qwen2.5-7B-Instruct (7B) , we use prompt template in official evaluation code repository of Qwen2.5-Math \footnote{https://github.com/QwenLM/Qwen2.5-Math}.
And for the experiments on Deepseek-R1-Distill-Qwen-1.5B (1.5B-R1), we use prompt template used in the training of Deepseek-R1 \citep{guo2025deepseek}.
The prompt templates are shown on Table \ref{tab:prompt_temp}.
We made slight modifications to the prompt for R1 to facilitate the extraction of the final answer from the \textbackslash boxed.

\begin{table*}[ht]
\centering
\caption{Prompt templates used in CROPI experiments. "<prompt>" will be replaced by specific training prompt during training. }
\label{tab:prompt_temp}
\begin{tabular}{p{0.2\linewidth}|p{0.8\linewidth}}
\toprule
\textbf{Setting} & \textbf{Prompt Templates} \\
\midrule
1.5B \& 7B  & \parbox[t]{0.8\linewidth}{
System: Please reason step by step, and put your final answer within \textbackslash boxed\{\}. \\
User: <prompt>
} \\
\midrule
1.5B-R1 & \parbox[t]{0.8\linewidth}{
System: A conversation between User and Assistant. The user asks a question, and the Assistant solves it.
The assistant first thinks about the reasoning process in the mind and then provides the user
with the answer. The reasoning process and answer are enclosed within <think> </think> and
<answer> </answer> tags, respectively, i.e., <think> reasoning process here </think>
<answer> answer here, and put your final answer within \textbackslash boxed\{\} </answer>.\\
User: <prompt>
} \\
\bottomrule
\end{tabular}
\end{table*}



\subsection{Hyperparameters of RL}
\label{sec:param_rl}

We present the key hyperparameters used for GRPO training on Table \ref{tab:hyperparameters}. 
For 1.5B / 7B / 1.5B-R1 experiment settings, we use max response length of \texttt{2048 / 4096 / 8192} respectively. 

\begin{table*}[ht]
\centering
\caption{Key Hyperparameters for GRPO Training}
\label{tab:hyperparameters}
\resizebox{\textwidth}{!}{%
\begin{tabular}{@{}lp{4cm}l@{}}
\toprule
\textbf{Parameter} & \textbf{Value} & \textbf{Description} \\ \midrule
\multicolumn{3}{l}{\textbf{RL Algorithm}} \\
\cmidrule(r){1-3}
Base Model &  Qwen2.5-1.5B-Instruct / Qwen2.5-7B-Instruct / DeepSeek-R1-Distill-Qwen-1.5B  & The pretrained model used as the starting point. \\
Batch Size & 128 & The number of prompts processed in each training step. \\
Max Prompt Length & 2048 & Maximum token length for the input prompt. \\
Max Response Length & 2048 / 4096 / 8192 & Maximum token length for the generated response. \\
Advantage Estimator & \texttt{grpo} & The specific RL algorithm used for training. \\
Actor Learning Rate & $1 \times 10^{-6}$ & The learning rate for the actor model's optimizer. \\
PPO Mini-batch Size & 128 & The batch size used for each PPO optimization update. \\
KL Regularization & True & Enables KL-divergence penalty against the reference model. \\
KL Coefficient ($\beta$) & 0.001 & Weight of the KL-divergence term in the loss function. \\
KL Loss Type & \texttt{low\_var\_kl} & A specific variant of KL loss calculation. \\
Entropy Coefficient & 0.001 & Weight of the entropy bonus to encourage exploration. \\
\addlinespace

\multicolumn{3}{l}{\textbf{Rollout \& Generation}} \\
\cmidrule(r){1-3}
Rollout Engine & \texttt{vllm} & The inference engine used for generating samples. \\
Samples per Prompt ($n$) & 8 & Number of candidate responses generated for each prompt. \\
Tensor Parallel Size & 4 & Degree of tensor model parallelism for rollouts. \\
Dynamic Batching & True & Enables dynamic batching for rollouts. \\
Max Batched Tokens & 16384 & Maximum number of tokens in a dynamic vLLM batch. \\
\addlinespace

\bottomrule
\end{tabular}%
}
\end{table*}

\section{Additional Analysis on Rollout Cost}
\label{sec:appendix_rollout_cost}

To clarify why rollout-free influence estimation is important in RLVR, we compare the cost of a rollout step against a single forward+backward step in the 1.5B setting. For a fair comparison, both settings use batch size 128 with sequence length capped at 8192. Table \ref{tab:rollout_cost} shows that rollout generation is substantially more expensive than gradient computation under comparable settings.

\begin{table}[ht]
\centering
\small
\caption{Cost comparison between rollout and forward+backward in the 1.5B setting.}
\label{tab:rollout_cost}
\begin{tabular}{lcc}
\toprule
Operation & Avg. Time (s) & Relative Cost \\
\midrule
Forward+Backward & 35.91 & 1.00x \\
Rollout & 319.80 & 8.91x \\
\bottomrule
\end{tabular}
\end{table}

This difference is also aligned with the computational characteristics of the two operations. A forward+backward step mainly consists of dense tensor computation and is typically compute-bound, while autoregressive rollout repeatedly reads model weights during decoding and is often constrained by memory bandwidth. As a result, methods that require fresh rollouts for every candidate prompt incur a much larger selection overhead than methods that reuse offline trajectories.

\section{Additional Analysis on Off-policy Gradient Estimator}
\label{sec:off_policy_estimator}
\begin{figure}[ht]
    \centering
    \includegraphics[width=0.50\textwidth]{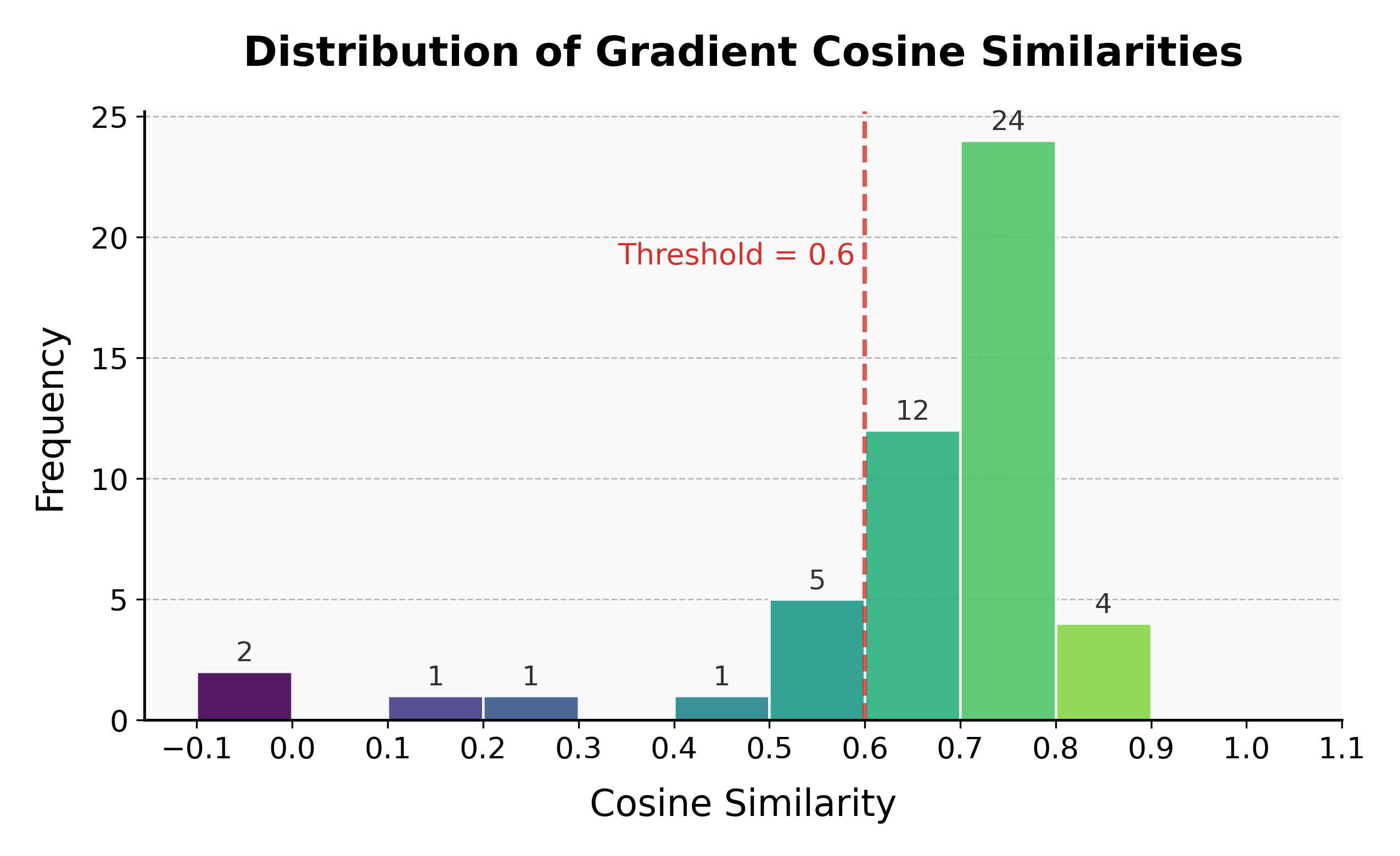}
    \caption{Cosine similarity distribution between on-policy gradients and off-policy gradients produced by our proposed estimator}
    \label{fig:cosine_similarity_distribution}
\end{figure}

We conduct supplementary experiments to  evaluate our proposed off-policy gradient estimator in \ref{eq:op_grad}. To demonstrate this point, we aim to verify that the off-policy gradient estimator produces gradients that are sufficiently consistent with the on-policy gradients at the step 200 checkpoint of CROPI. Specifically, we collect $50$ problems on which both the original Qwen2.5-1.5B-Instruct model and the step 200 checkpoint of CROPI neither achieve perfect success nor complete failure across $8$ rollouts. Using Equation \ref{eq:grpo}, we obtain the on-policy gradients for these $50$ problems, and using Equation \ref{eq:op_grad}, we compute the corresponding off-policy estimated gradients. We then calculate the cosine similarity between each pair of on-policy and off-policy gradients and visualize the distribution in Figure \ref{fig:cosine_similarity_distribution}. As shown, $40$ out of the $50$ pairs exhibit a cosine similarity greater than $0.6$, which provides strong evidence for the effectiveness of our proposed off-policy gradient estimator.
We further investigate the rank preservation capability of the off-policy gradient estimator. Specifically, based on the two sets of $50$ gradients obtained above, we construct two $50 \times 50$ cosine similarity matrices and compute the index ranking for each row. In the ideal case, the ranking orders of corresponding rows in the two matrices should be identical, indicating perfect rank consistency of the estimator. However, when evaluating the top 10\% rank preservation across all gradients of interest, the off-policy gradient estimator achieves only 28.80\% consistency. Although this performance is significantly higher than that of a random baseline, we believe there remains substantial room for improving rank preservation.

\section{Additional Analysis on Gradient Projection}
\label{sec:add_ana_proj}
In order to analyze the rank preservation efficiency of gradient projection method we used, we provide additional experimental results and one possible explanation for the outcomes.

Following the setting of Section \ref{sec:ana_random_projection}, we analyze the influence of the sparse ratio for the 3B model. We observe the same pattern of rank preservation curve as in the 1.5B model experiment. To be specific, we observe that when we use the full gradient before applying the random projection, the \texttt{precision@10\%} is only slightly above the random guess. However, when the sparse ratio reaches 0.1, the \texttt{precision@10\%} increases to more than 80\%.

\begin{figure}[ht]
    \centering
    \includegraphics[width=0.50\textwidth]{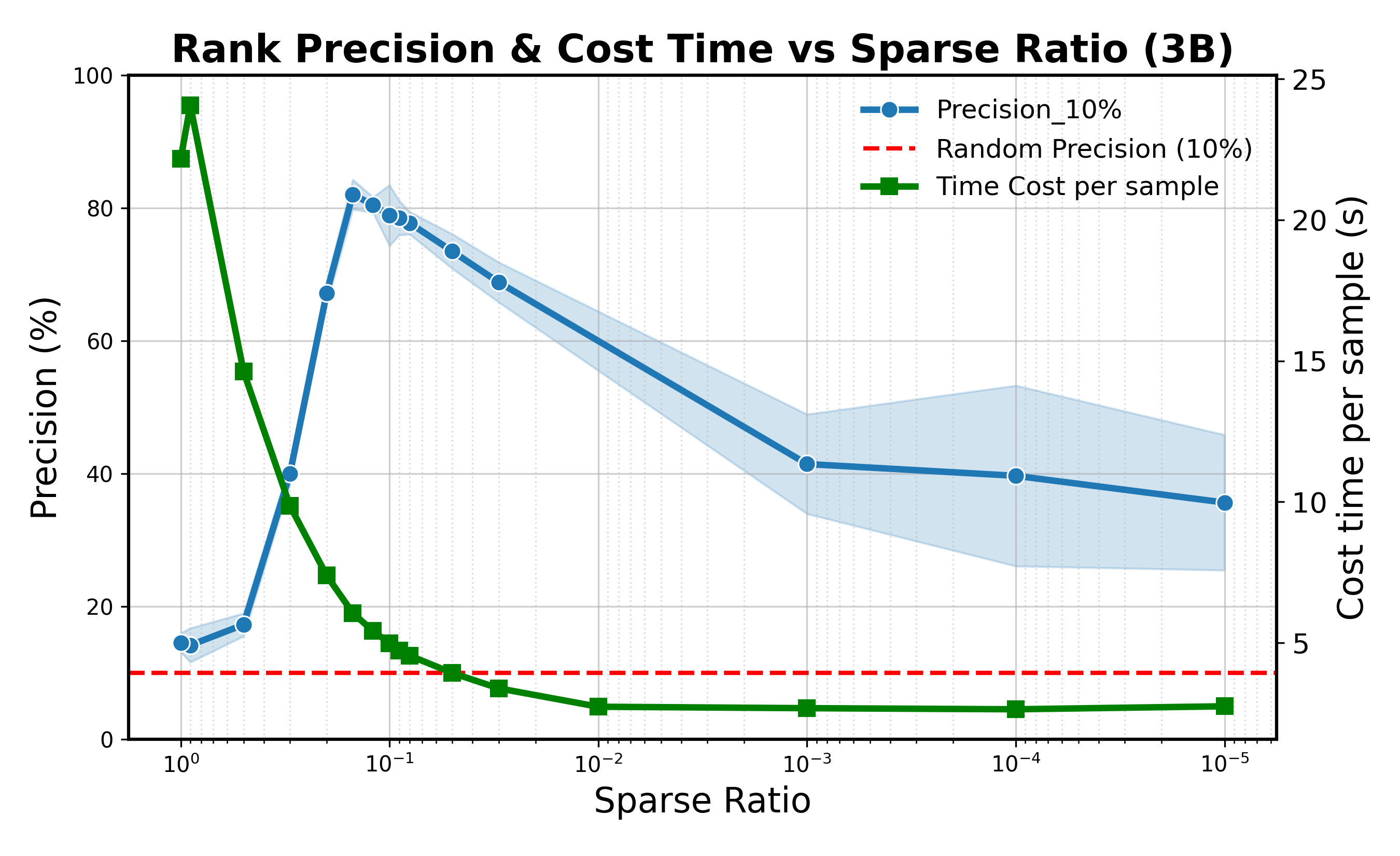}
    \caption{Rank preservation experiments for Sparse Random Projection under 3B model setting.}
    \label{fig:proj_rank_preservation_3B}
\end{figure}

\begin{figure}[ht]
    \centering
    \includegraphics[width=0.50\textwidth]{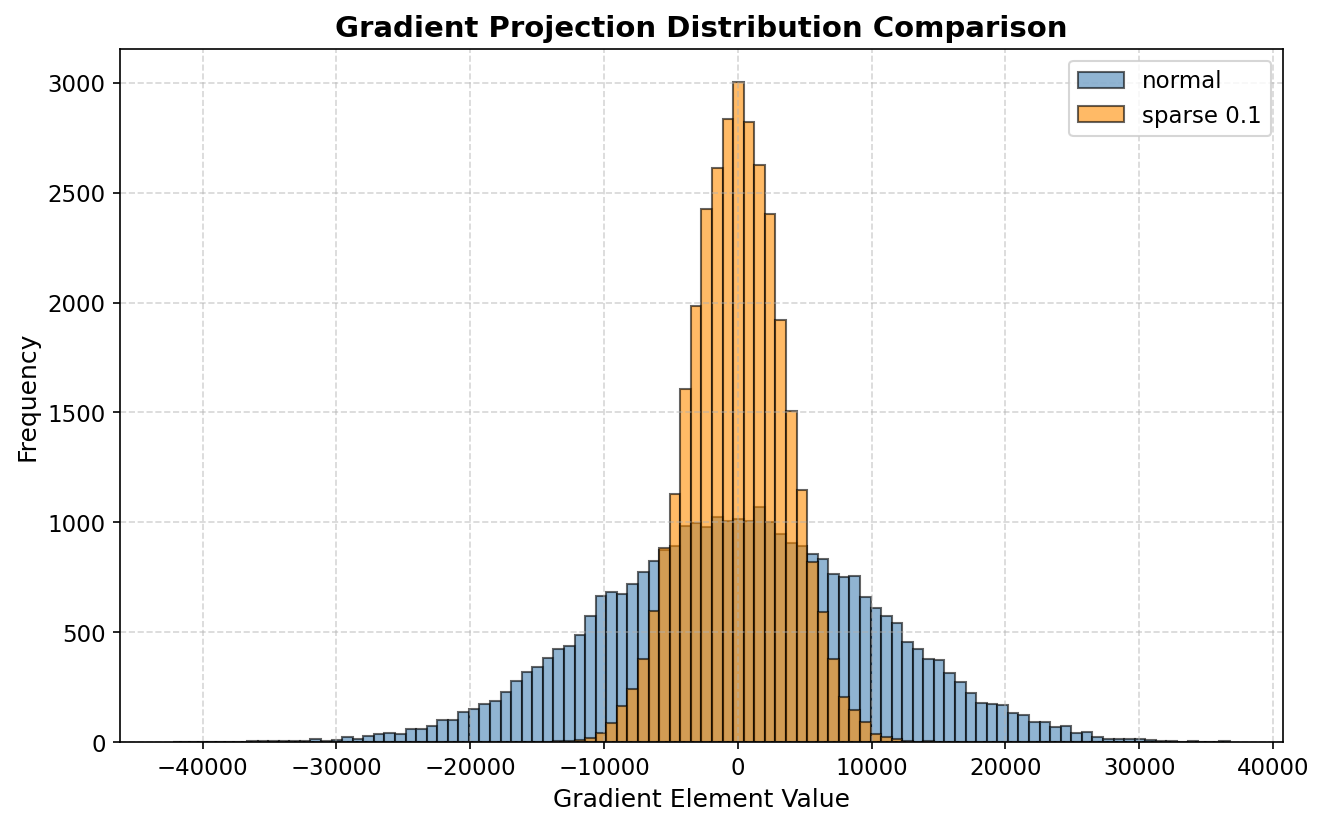}
    \caption{Element values distribution of gradients produced by normal random projection and sparse random projection with a sparse ratio of 0.1.}
    \label{fig:element_distribution_compare}
\end{figure}

As mentioned in \ref{sec:ana_random_projection}, we hypothesize that sparsification reduce numerical error in projection operation, which is conducted in float 16 format. Serious numerical instability and precision loss can happen during projection. Sparsity might help filter out much of the error by masking a bunch of information. In Figure \ref{fig:element_distribution_compare}, we plot the histogram of the element value distribution of the projected gradients. We observe that conducting random projection after sparsification cluster more element values towards zero, which might help avoid precision loss in rank preservation. We leave further exploration of this area to future work.

\section{Additional Analysis on Data Selected by CROPI}
\label{sec:more_select}

Following the setting of Section \ref{sec:ana_data}, we put more analysis results in this section. 

\subsection{Data Source of Selected Prompts}
\begin{figure}[ht]
    \centering
    \includegraphics[width=0.50\textwidth]{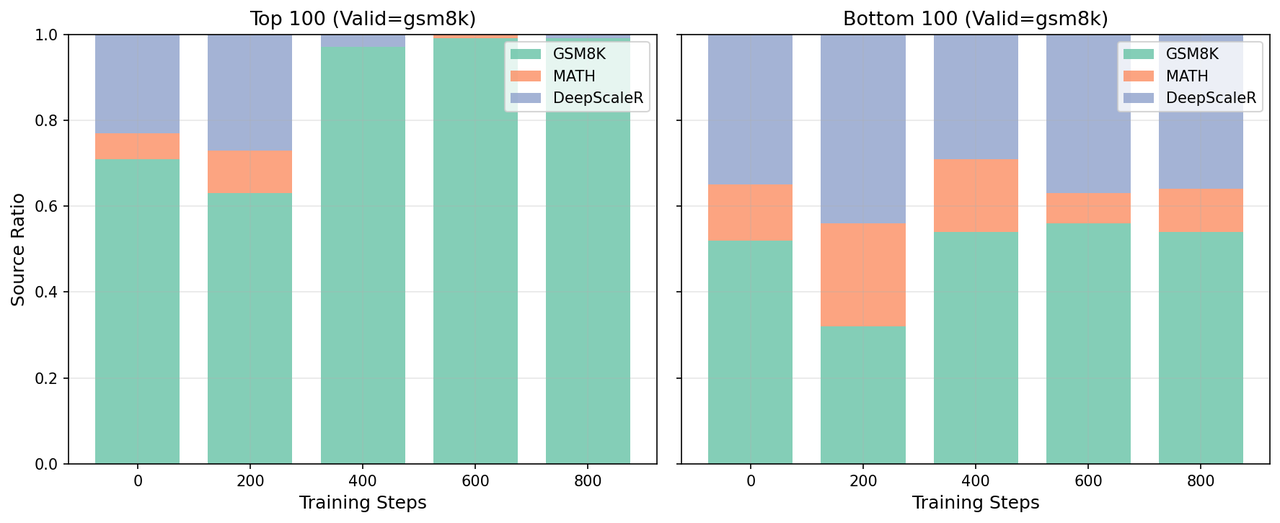}
    \caption{The source dataset of top-100 and bottom-100 training prompts selected by POPI with validation set GSM8K in different training steps. }
    \label{fig:cropi_gsm_datasrc}
\end{figure}

\begin{figure}[ht]
    \centering
    \includegraphics[width=0.50\textwidth]{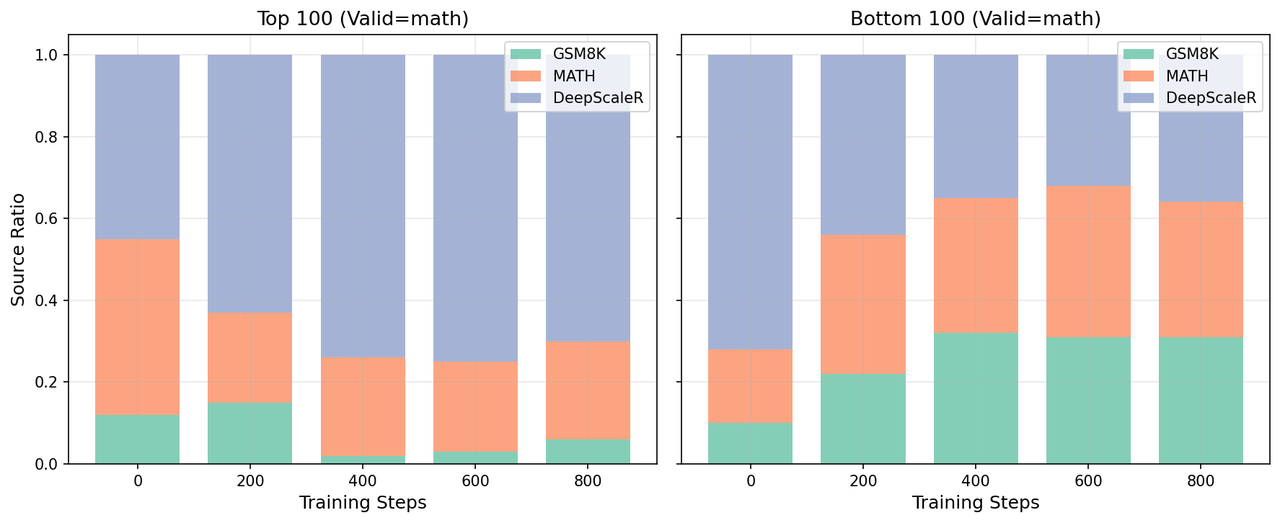}
    \caption{The source dataset of top-100 and bottom-100 training prompts selected by POPI with validation set MATH in different training steps. }
    \label{fig:cropi_math_datasrc}
\end{figure}

We first analyzed the dataset origins of the top 100 training examples selected using POPI with different validation sets, as shown in Figure \ref{fig:cropi_gsm_datasrc} and Figure \ref{fig:cropi_math_datasrc}.
Across multiple rounds of data selection, we observe a pronounced shift in the composition of training examples deemed most relevant for the GSM8K validation set. Initially, in Round 0, 71\% of the top 100 selected examples are sourced from GSM8K data and 6\% from MATH.
However, by Round 4 (step=800), GSM8K’s representation rises to 99\%, while MATH data are almost entirely excluded. This dynamic indicates that the selection algorithm increasingly prioritizes GSM8K training samples that are most advantageous for improving performance on the GSM8K validation set. 
Furthermore, the process simultaneously eliminates GSM8K examples that are particularly ineffective for the MATH validation set, underscoring a nuanced mechanism of data curation that both refines selection towards task relevance and discards samples detrimental to cross-task generalization.

\subsection{Knowledge Domains of Selected Prompts}

\begin{figure*}[htbp]
    \centering
    \includegraphics[width=0.80\textwidth]{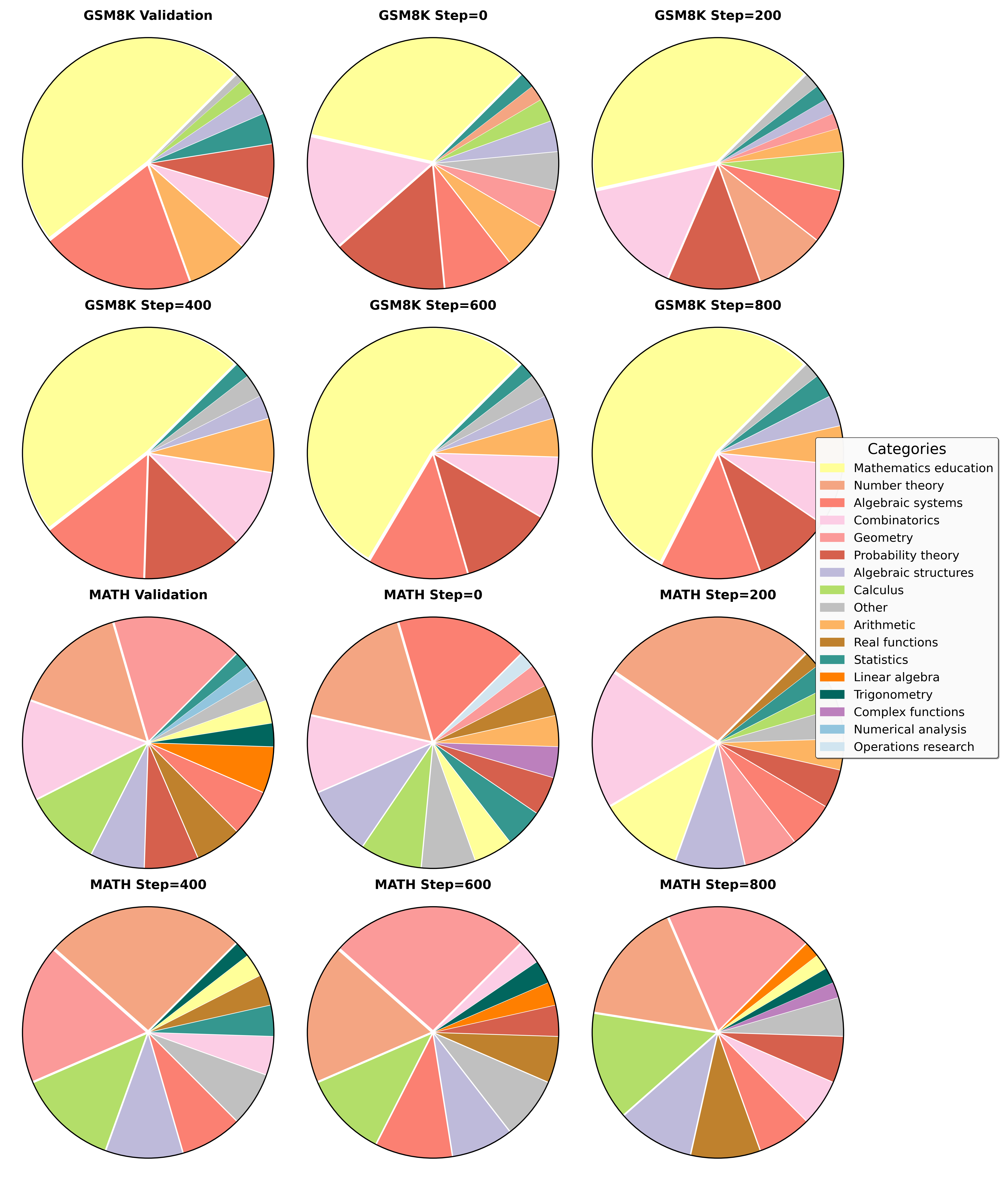}
    \caption{The knowledge dimains of top-100 and training prompts selected by POPI with validation set MATH in different training steps. 
    "Validation" denotes the knowledge category distribution of the validation set, while "step=X" refers to the distribution of knowledge categories within the top-100 selected samples at the checkpoint corresponding to step=X.}
    \label{fig:cropi_sel_catagories}
\end{figure*}

Since direct inspection of each individual problem makes it difficult to observe distinctive patterns, we adopted the categorization approach from s1 \cite{muennighoff2025s1simpletesttimescaling}. Utilizing \texttt{GPT-4o}, we classified both the validation set (randomly sampled 100 questions) and the top-100 mathematics problems selected by POPI according to the Mathematics Subject Classification (MSC) system (e.g., geometry, combinatorics, etc.) from the American Mathematical Society\footnote{https://mathscinet.ams.org/mathscinet/msc/msc2020.html}. 
The results (Figure \ref{fig:cropi_sel_catagories}) show that the distribution of knowledge domains in the selected data closely mirrors that of the validation set itself. 
Moreover, as training progresses, the distribution of knowledge domains among the selected samples also shifts dynamically to meet the need of online policy.

\begin{figure}[ht]
    \centering
    \includegraphics[width=0.50\textwidth]{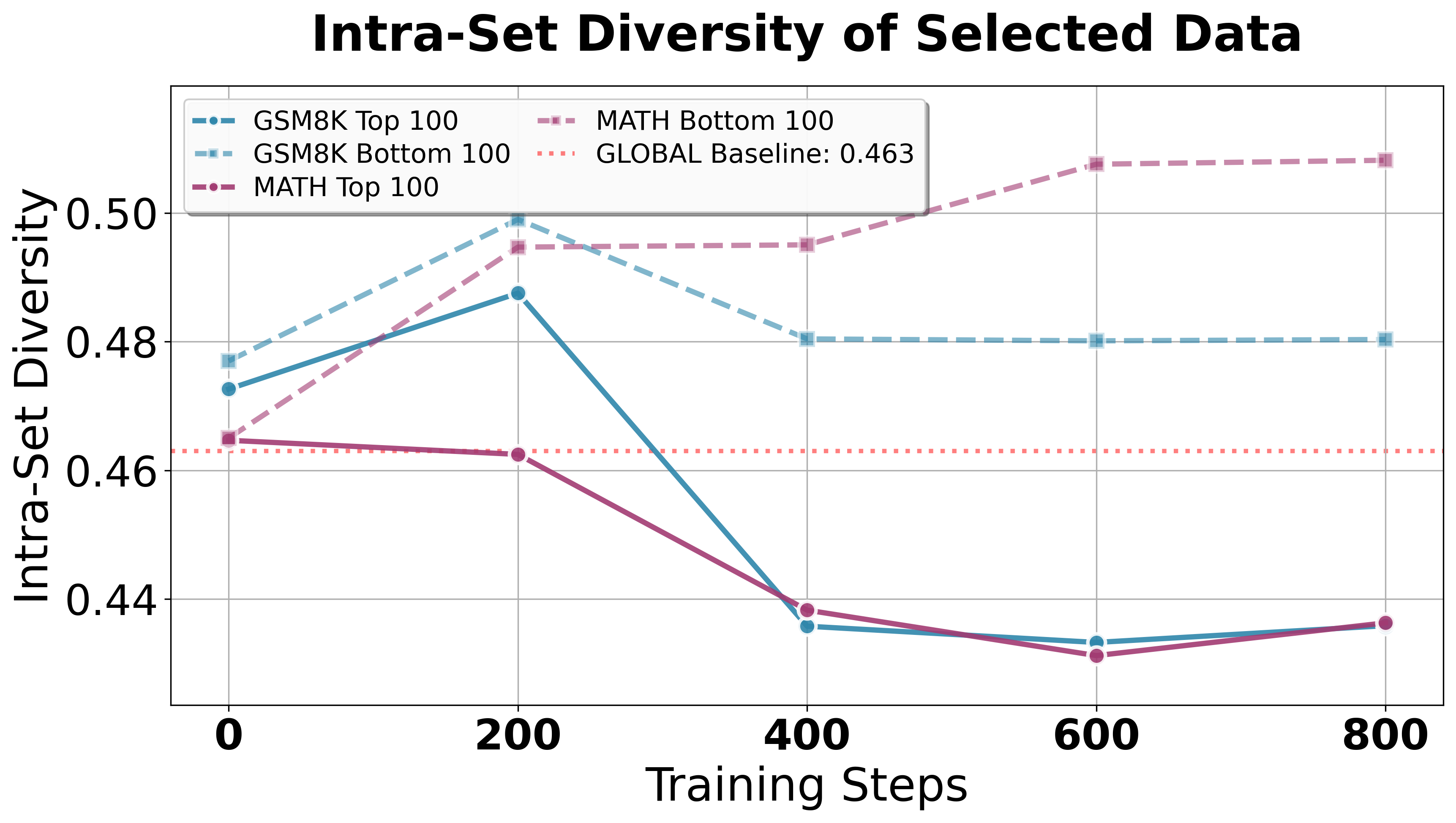}
    \caption{The semantic diversity of top-100 and bottom-100 training prompts selected by POPI with validation set GSM8K and MATH in different training steps.}
    \label{fig:cropi_sel_diversity}
\end{figure}

\subsection{Semantic Diversity of Selected Prompts}

We also evaluated the internal diversity of the top-100 and bottom-100 training samples selected by POPI, quantified by $1 - \mathbb{E}_{e_i, e_j \in E}[\texttt{cossim}(e_i, e_j)]$, where $E$ denotes the set of semantic embeddings for each dataset. 
As shown in Figure \ref{fig:cropi_sel_diversity}, the diversity of the POPI-selected top-100 samples is noticeably lower compared to both the random baseline and the bottom-100 set. 
This indicates that training examples highly relevant to the validation set tend to be semantically similar, which aligns with our intuition. 
Such reduced diversity could potentially limit the generalization ability of the trained model. 
Employing multiple diverse validation sets and aggregating the selected training samples from each can effectively mitigate this issue. 
Our experimental results further demonstrate that models trained with CROPI exhibit performance gains even on untargeted test sets, indicating good generalization capability of the CROPI approach.

\subsection{Human Annotated Difficulty-level of Selected Prompts}

\begin{figure}[ht]
    \centering
    \includegraphics[width=0.50\textwidth]{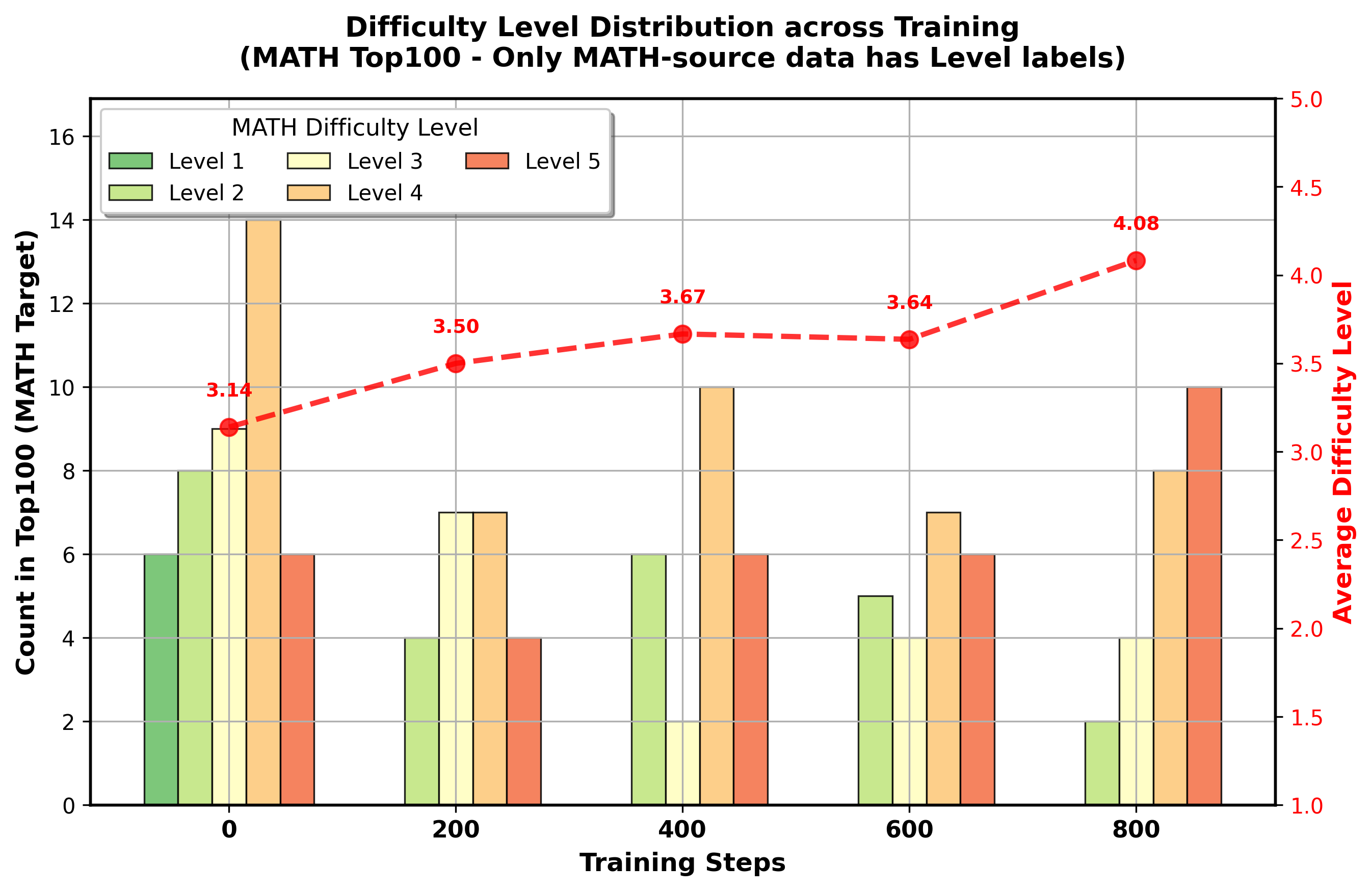}
    \caption{The human annotated difficulty-levels of top 100 prompts in MATH-Train dataset}
    \label{fig:cropi_math_level}
\end{figure}

Since MATH includes human-annotated difficulty levels, we also recorded the difficulty levels of MATH-Train samples selected when MATH served as the validation set, as depicted in Figure \ref{fig:cropi_math_level}. 
It can be observed that, as training progresses, CROPI increasingly favors samples with higher difficulty levels. This trend suggests a continual expansion of the model's capability boundaries, requiring the selection of progressively more challenging problems.

\subsection{Most influential questions}

\begin{figure}[htbp]
    \centering
    \includegraphics[width=0.50\textwidth]{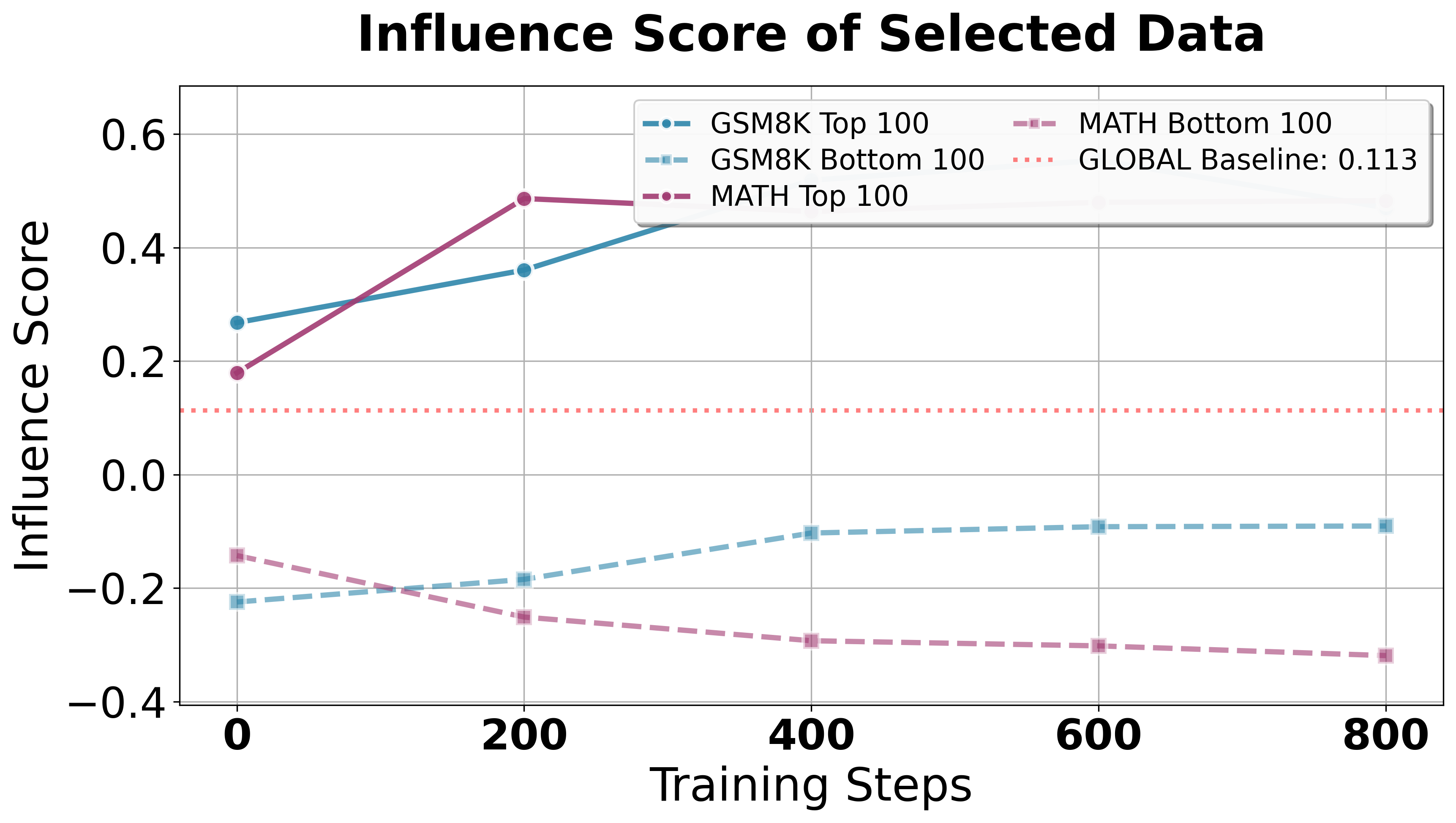}
    \caption{The influence scores of top-100 and bottom-100 training prompts selected by POPI with validation set GSM8K and MATH in different training steps.}
    \label{fig:cropi_sel_inf}
\end{figure}

We also visualized the influence scores of the top-100 and bottom-100 training samples calculated by POPI across different rounds, as shown in Figure \ref{fig:cropi_sel_inf}. 
The influence scores for the top-100 samples exhibit a steadily increasing trend, eventually stabilizing around 0.5. In contrast, the bottom-100 samples display substantial divergence from the validation set in gradient space, with cosine similarity values falling below zero. 
We put the most infuential prompts selected by POPI with MATH validation set MATH in Table \ref{tab:most_inf}, which reflects the training dynamics of the model in a data perspective.

\begin{table*}[htbp]
\centering
\begin{tabular}{cccp{7cm}}
\toprule
\textbf{Round} & \textbf{Rank} & \textbf{POPI Score} & \textbf{Prompt} \\
\hline
0 & 1 & 0.2166 & Let $m$ and $n$ satisfy $mn=4$ and $m+n=5$. What is $|m-n|$? \\
\hline
0 & 2 & 0.2160 & If $x+2y-3z=7$ and $2x-y+2z=6$, determine $8x+y$. \\
\hline
0 & 3 & 0.2136 & If $\sqrt{5 + x} + \sqrt{20 - x} = 7$, what is the value of $(5 + x)(20 - x)$? \\
\midrule
1 & 1 & 0.5534 & A bird discovered $543_{8}$ different ways to build a nest in each of its eight tree homes. How many ways are there in base 10? \\
\hline
1 & 2 & 0.5479 & How many three-eighths are there in $8\frac{5}{3} - 3$? \\
\hline
1 & 3 & 0.5434 & A secret facility is a rectangle measuring $200 \times 300$ meters... How many meters did the fourth guard run to reach the intruder? \\
\midrule
2 & 1 & 0.5382 & Determine the least possible value of $(x+2)(x+3)(x+4)(x+5)+2024$ where $x$ is a real number.  \\
\hline
2 & 2 & 0.5314 & What is the total volume and the total surface area in square feet of three cubic boxes if their edge lengths are 3 feet, 5 feet, and 6 feet, respectively? \\
\hline
2 & 3 & 0.5263 & Find the sum of $543_7$, $65_7$, and $6_7$ in base $7$. \\
\midrule
3 & 1 & 0.5716 & Suppose that $x$ and $y$ are positive numbers with $xy=\frac{1}{9}$, $x(y+1)=\frac{7}{9}$, and $y(x+1)=\frac{5}{18}$... What is the value of $(x+1)(y+1)$? \\
\hline
3 & 2 & 0.5623 & Given the plane vectors $\overrightarrow{a}$ and $\overrightarrow{b}$... calculate the angle between vectors $\overrightarrow{a}$ and $\overrightarrow{b}$. \\
\hline
3 & 3 & 0.5457 & What is the smallest positive integer $n$ such that $5n \equiv 105 \pmod{24}$? \\
\midrule
4 & 1 & 0.5555 & In a set of 15 different-colored markers, how many ways can Jane select five markers if the order of selection does not matter? \\
\hline
4 & 2 & 0.5532 & What is the greatest common divisor of $1729$ and $1768$? \\
\hline
4 & 3 & 0.5472 & For what value of $n$ does $|6 + ni| = 6\sqrt{5}$? \\
\bottomrule
\end{tabular}
\caption{Most Influential prompts in different rounds. Round 0 means the first round of data selection, corresponding to training step of 0. Round 1 corresponds to training step 200, and so on.}
\label{tab:most_inf}
\end{table*}

\section{Additional Results on CROPI Experiments}
\label{sec:detail_rl}



\begin{figure*}[htbp]
    \centering
    \includegraphics[width=1.0\textwidth]{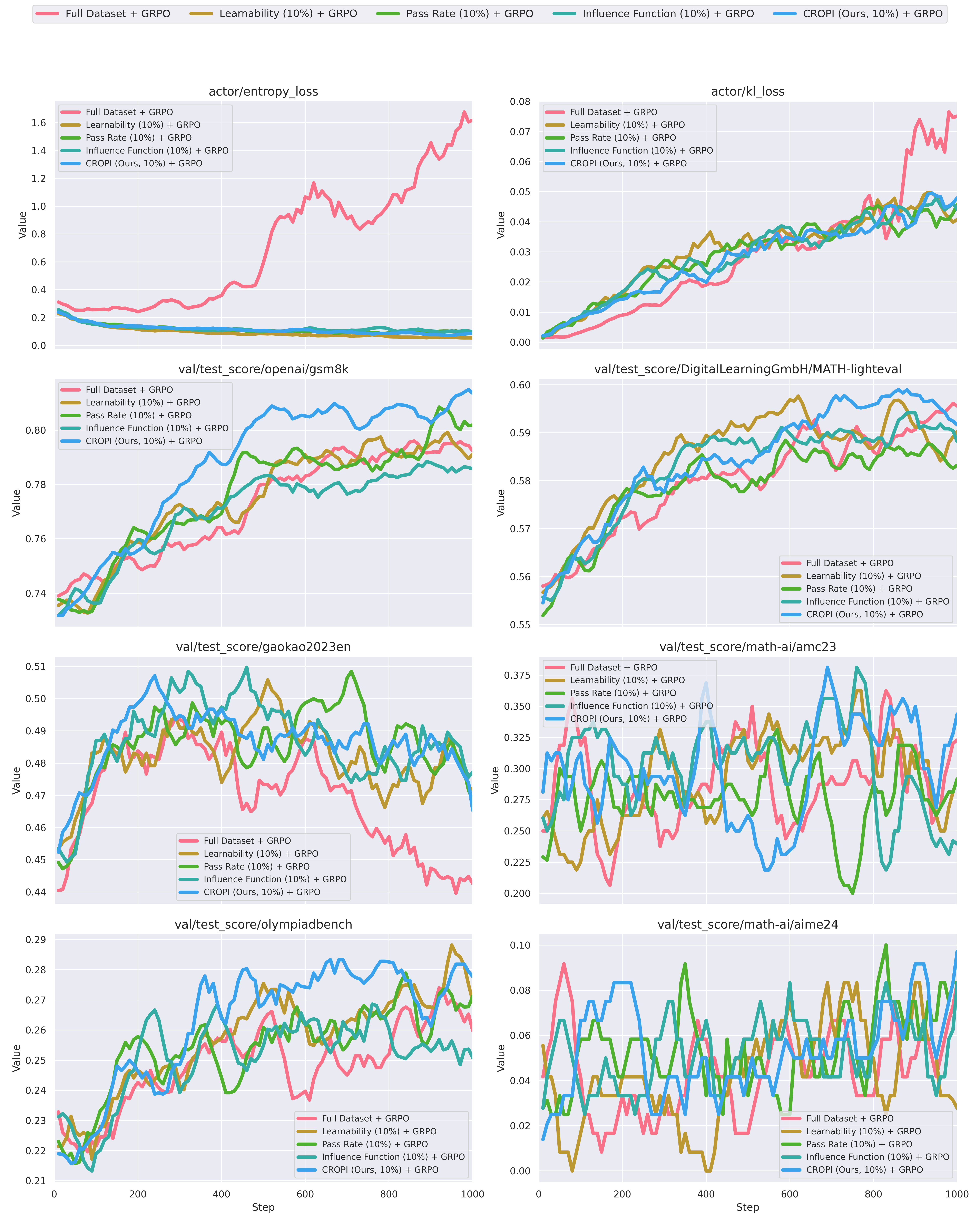}
    \caption{Training curves in 1.5B setting: KL Loss. Entropy Loss,  Accuracy across different test sets. }
    \label{fig:1.5b_curve_full}
\end{figure*}

\begin{figure*}[htbp]
    \centering
    \includegraphics[width=1.0\textwidth]{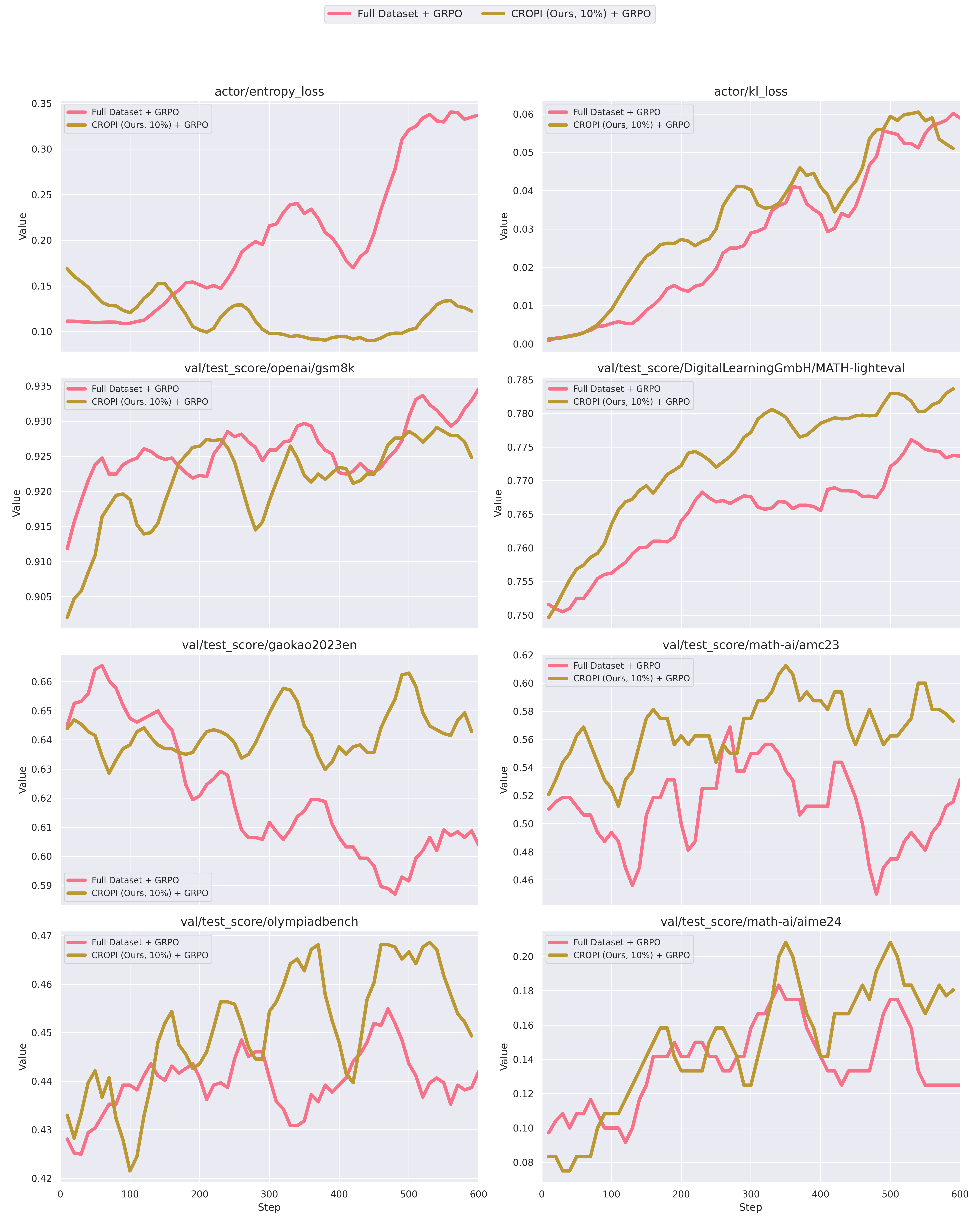}
    \caption{Training curves in 7B setting: KL Loss. Entropy Loss,  Accuracy across different test sets. 
    }
    \label{fig:7b_curve_full}
\end{figure*}

\begin{figure*}[htbp]
    \centering
    \includegraphics[width=1.0\textwidth]{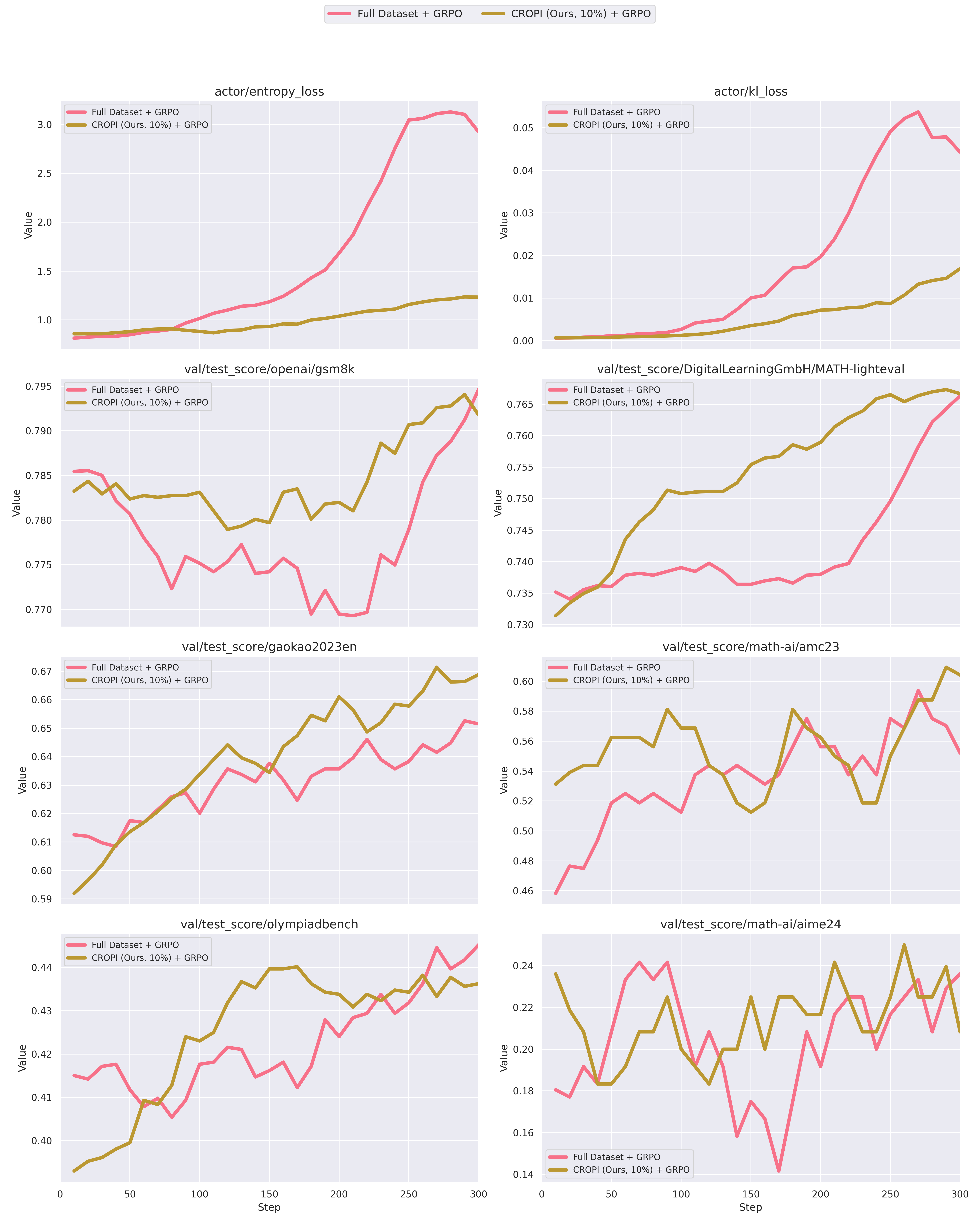}
    \caption{Training curves in 1.5B-R1 setting: KL Loss. Entropy Loss,  Accuracy across different test sets. 
    }
    \label{fig:1.5b_r1_curve_full}
\end{figure*}

We plot the main recorded metrics from RL training under different settings in Figure \ref{fig:1.5b_curve_full}, \ref{fig:7b_curve_full}, \ref{fig:1.5b_r1_curve_full}. 
As illustrated in these figures, the entropy of the data selection methods is significantly lower than that of the full data baseline during RL training, which may hinder policy exploration in subsequent training stages.

\end{document}